\documentclass[review]{elsarticle}

\usepackage{hyperref}

\usepackage{amsmath}

\usepackage{xcolor}

\usepackage{booktabs,dcolumn}

\usepackage{graphicx,multirow,bm}
\usepackage{threeparttable}

\usepackage{algorithm}

\usepackage{algorithmic}
\journal{Journal of \LaTeX\ Templates}

\begin{document}

\begin{frontmatter}

\title{Divide-and-Conquer Large Scale Capacitated Arc Routing Problems with Route Cutting Off Decomposition}





\author[mymainaddress]{Yuzhou Zhang}
\author[mysecondaryaddress]{Yi Mei\corref{mycorrespondingauthor}}
\cortext[mycorrespondingauthor]{Corresponding author}
\ead{yi.mei@ecs.vuw.ac.nz}
\author[mymainaddress]{Buzhong Zhang}
\author[mymainaddress]{Keqin Jiang}

\address[mymainaddress]{School of Computer and Information, Anqing Normal University, Anqing 246133, China}
\address[mysecondaryaddress]{School of Engineering and Computer Science, Victoria University of Wellington, Kelburn 6012, New Zealand}

\begin{abstract}
The capacitated arc routing problem is a very important problem with many practical applications. This paper focuses on the large scale capacitated arc routing problem. Traditional solution optimization approaches usually fail because of their poor scalability. The divide-and-conquer strategy has achieved great success in solving large scale optimization problems by decomposing the original large problem into smaller sub-problems and solving them separately. For arc routing, a commonly used divide-and-conquer strategy is to divide the tasks into subsets, and then solve the sub-problems induced by the task subsets separately. However, the success of a divide-and-conquer strategy relies on a proper task division, which is non-trivial due to the complex interactions between the tasks. This paper proposes a novel problem decomposition operator, named the route cutting off operator, which considers the interactions between the tasks in a sophisticated way. To examine the effectiveness of the route cutting off operator, we integrate it with two state-of-the-art divide-and-conquer algorithms, and compared with the original counterparts on a wide range of benchmark instances. The results show that the route cutting off operator can improve the effectiveness of the decomposition, and lead to significantly better results especially when the problem size is very large and the time budget is very tight.
\end{abstract}

\begin{keyword}
\texttt{}Capacitated arc routing problem\sep route cutting off\sep large scale optimization\sep divide-and-conquer
\end{keyword}

\end{frontmatter}

\normalsize \normalsize \abovedisplayskip=2.0pt plus 2.0pt minus 2.0pt \belowdisplayskip=2.0pt plus 2.0pt minus 2.0pt \baselineskip 16pt

\section{Introduction}
\label{sec:intr}

The Capacitated Arc Routing Problem (CARP) is a very important combinatorial optimization problem with a wide range of real-world applications such as winter gritting \cite{handa2006robustmagz}, mail delivery \cite{eiselt1995arc}, urban waste collection \cite{mei2011memetic,zhang2017memetic,chen2018phased}, and snow removal \cite{polacek2008variable}. First presented by Golden and Wong \cite{golden1981capacitated}, it seeks an optimal set of routes (e.g. cycles starting and ending at a depot) for a fleet of vehicles to serve the edges in a graph subject to certain constraints. There have been extensive studies for solving CARP, and a large number of competitive approaches have been proposed, e.g. \cite{ulusoy1985fleet,beullens2003guided,lacomme2004competitive,longo2006solving,tang2009memetic,feng2010towards,mei2011decomposition}.

In real world, the problem size of CARP can usually be very large (e.g. thousands of streets in a city need to be served for waste collection). It has been demonstrated that most existing approaches exploring the entire search space of the original problem have poor scalability, i.e. their effectiveness deteriorate rapidly as the problem size grows \cite{brandao2008deterministic,martinelli2011improved,mei2009global}. As a result, most approaches that are competitive for the small and medium instances cannot solve Large Scale CARP (LSCARP) effectively.

The divide-and-conquer strategy is a promising approach to address the scalability issue of algorithms when solving large scale optimization problems. The main idea is to decompose the original large problem into smaller sub-problems that can be solved individually. There have been several divide-and-conquer approaches proposed for solving LSCARP, such as the Route Distance Grouping (RDG) decomposition \cite{mei2014cooperative} and Hierarchical Decomposition (HD) \cite{tang2017scalable}, which have achieved great success in finding competitive solutions efficiently. The main idea of these approaches is to divide the tasks (i.e. required edges) into subsets, solve the sub-problems induced by the subsets separately, and finally combine the solutions to the sub-problems (subsets of routes) together to form the solution to the original problem.

For designing divide-and-conquer approaches for LSCARP, the problem decomposition is the key step. In an \emph{ideal} decomposition, the subsets of tasks are completely independent of each other, and the combination of the optimal solutions to the corresponding sub-problems (union of the routes) can result in the optimal solution to the original problem. However, it is very challenging to identify the ideal decomposition due to the complex interactions between the tasks, if not impossible. Therefore, all the existing divide-and-conquer approaches (e.g. \cite{mei2014cooperative,shang2016improved,shang2017quantum-Inspired,shang2017memetic,tang2017scalable}) adopt the \emph{adaptive} decomposition. They start with randomly or heuristically generated task subsets, and gradually improve the decomposition during the search process based on the updated information.


Intuitively, the quality of a decomposition depends on two major factors. First, the tasks belonging to the same route in the optimal solution should be in the same subset to prevent from breaking an optimal route. Second, the tasks that are close to each other should be more likely to be in the same subset, so that solving the corresponding sub-problem can lead to better solutions (i.e. shorter routes). It is challenging to consider these two factors simultaneously. Although there have been some effort from existing approaches \cite{mei2014cooperative,tang2017scalable}, there is still great potential for improvement.

The goal of this paper is to propose a new decomposition scheme to consider the above two factors in a sophisticated way during the search process. More specifically,
\begin{itemize}
    \item we design a \emph{task rank matrix} according to the distances between different tasks and define \emph{good links} and \emph{poor links} between tasks based on the task rank matrix;
    \item we develop a new Route Cutting Off (RCO) decomposition operator, which is more likely to decompose a route by breaking poor links rather than good links;
    \item we develop two new divide-and-conquer algorithms by embedding the RCO operator into two state-of-the-art algorithms (i.e. RDG-MAENS \cite{mei2014cooperative} and SAHiD \cite{tang2017scalable});
    \item we verify the effectiveness of the proposed RCO operator by comparing the newly developed algorithms with their original counterpart and other state-of-the-art approaches on a range of LSCARP instances.
\end{itemize}

The remainder of this paper is organized as follows. First, Section \ref{sec:back} gives the background, including the problem description and related work. Then, the proposed RCO operator is introduced in Section \ref{sec:RCO}. Section \ref{sec:exp} presents the experimental studies and discussions on the results. Finally, the conclusions and future work are given in Section \ref{sec:conc}.

\section{Background}
\label{sec:back}

In this section, we briefly describe the problem statement and related work.

\subsection{Capacitated Arc Routing Problem}
\label{sec:CARP}

Given an undirected graph $G(V,E)$, where $V$ and $E$ represent the vertex and edge sets. Each edge $e \in E$ has a demand $d(e)\geq 0$, a non-negative service cost $sc(e)\geq 0$ and a non-negative deadheading cost $dc(e)\geq 0$. An edge with a positive demand is called a \emph{required edge} or a \emph{task}. The set of all the tasks is denoted as $T = \{e \in E | d(e) > 0\} \subseteq E$. A fleet of vehicles with a limited capacity $Q$ is located at the \emph{depot} $v_0 \in V$ to serve the tasks. CARP is to design a set of least-cost routes for the vehicles to serve all the tasks subject to the following constraints:
\begin{itemize}\itemsep 0pt
    \item each route starts and ends at the depot (i.e. is a cycle);
    \item each task is served exactly once by a vehicle;
    \item (\emph{capacity constraint}) the total demand served by each vehicle cannot exceed its capacity.
\end{itemize}

Under the task representation \cite{tang2009memetic}, each task $(u,v)$ can be represented by two IDs, each representing one of its directions. In addition to the demand, service cost and deadheading cost of the corresponding task, each ID $t$ is associated with a \emph{head vertex} $hv(t)$, a \emph{tail vertex} $tv(t)$ and a inverse ID $inv(t)$. Specifically, for task $(u,v)$ and its two IDs $t_1$ and $t_2$, we have $hv(t_1) = tv(t_2) = u$, $tv(t_1) = hv(t_2) = v$, $t_2 = inv(t_1)$ and $t_1 = inv(t_2)$.

A CARP solution $S$ can be represented as a set of routes, i.e. $S = \{S_1, \dots, S_{|S|}\}$, where $S_k$ ($k=1,\dots,|S|$) is the $k$th route. Each route $S_k$ is represented as a sequence of task IDs $S_k = (t_{k1}, \dots, t_{k|S_k|})$.

The total cost of $S$ is calculated as follows:
\begin{equation}
tc(S) = \sum_{k=1}^{|S|}\sum_{i = 1}^{|S_k|-1}\Big[sc(S_k[i])+\delta(tv(S_k[i]),hv(S_k[i+1]))\Big], \label{eq:tc}
\end{equation}
where $S_k[i]$ stands for the $i$th element (task ID) in $S_k$. $\delta(u,v)$ indicates the cost of the shortest path from vertices $u$ to $v$, which can be calculated by Dijkstra's algorithm \cite{dijkstra1959note} beforehand.

Then, CARP can be formulated as follows:
\begin{eqnarray}
    \min \;\;\; & tc(S),  \label{eq:obj} \\
    s.t. \;\;\; & S_k[1] = S_k[|S_k|] = t_0, \forall k = 1, \dots, |S|, \label{eq:con1} \\
    & \sum_{k = 1}^{|S|}(|S_k|-2) = |T|, \label{eq:con2} \\
    & S_{k}[i] \neq S_{k'}[i'], \;\; \forall \; 1 \leq k, k' \leq |S|, 2 \leq i, i' \leq |S_k|-1, k \neq k'  \mbox{ or }  i \neq i', \label{eq:con3} \\
    & S_{k}[i] \neq inv(S_{k'}[i']), \;\; \forall \; 1 \leq k, k' \leq |S|, 2 \leq i, i' \leq |S_k|-1, k \neq k'  \mbox{ or }  i \neq i', \label{eq:con4} \\
    & \sum_{i = 2}^{|S_k|-1}d(S_k[i]) \leq Q, \;\; \forall \;\; 1 \leq k \leq |S|, \label{eq:con5} \\
    & S_{k}[i] \in T, \;\; \forall \; k = 1, \dots, |S|, i = 2, \dots, |S_k|-1. \label{eq:domain}
\end{eqnarray}
where in Eq. (\ref{eq:con1}), $t_0$ is a dummy task ID representing the depot loop, i.e. $hv(t_0) = tv(t_0) = v_0$, $d(t_0) = sc(t_0) = dc(t_0) = 0$ and $inv(t_0) = t_0$. Eq. (\ref{eq:obj}) is the objective function, which is to minimize the total cost calculated by Eq. (\ref{eq:tc}). Eq. (\ref{eq:con1}) specifies that in each route $S_k$, the first and last element must be the depot loop, indicating that each route starts and ends at the depot. Eq. (\ref{eq:con2}) means that the total number of task IDs served by all the routes (excluding the first and last elements which are the depot loop) equals the total number of tasks. Eqs.  (\ref{eq:con3}) and (\ref{eq:con4}) implies that any two task IDs served at different positions of the routes belong to different tasks. Therefore, Eqs. (\ref{eq:con2})--(\ref{eq:con4}) guarantee that each task is served exactly once. Eq. (\ref{eq:con5}) indicates  that the total demand served by each route is no greater than its capacity. Eq. (\ref{eq:domain}) defines the domain of the elements of each route, i.e. except the first and last element, all the other elements must belong to the task set.


\subsection{Related Work}
\label{sec:rw}


Since presented in 1981 \cite{golden1981capacitated}, CARP has received a lot of research interests over the past decades, and a variety of competitive algorithms ranging from mathematical programming (e.g. \cite{belenguer2003cutting,baldacci2010exact,schlebusch2012cut,bartolini2013improved}) and heuristic solution search approaches (e.g. \cite{lacomme2004competitive,beullens2003guided,mourao2005heuristic,longo2006solving,tang2009memetic,feng2010towards,yao2017memetic,shang2016immune}) have been proposed for solving it. However, most of the early studies focused on small and medium scaled problems, and only tested on small and medium-sized benchmark instances. For example, the most commonly used gdb \cite{dearmon1981comparison}, val \cite{benavent1992capacitated}, egl \cite{eglese1996tabu} and Beullens' benchmark sets \cite{beullens2003guided} have no more than 190 tasks.

In 2008, Brand{\~a}o and Eglese \cite{brandao2008deterministic} generated a large scale dataset named EGL-G, in which the number of tasks was increased to 375. Since then, the research interests gradually shifted to the scalability of the approach, and more and more studies were conducted specifically to tackle LSCARP. For example, Brand{\~a}o and Eglese \cite{brandao2008deterministic} proposed a tabu search and achieved promising results on the EGL-G instances. Mei et al. \cite{mei2009global} proposed a tabu search with a global repair operator, and Martinelli et al. \cite{martinelli2011improved} proposed an Iterative Local search based on Random Variable Neighbourhood Descent (ILS-RVND), both of which improved the upper bounds of the EGL-G dataset. Vidal  \cite{vidal2017node} proposed an unified hybrid genetic search (UHGS) which outperformed all the existing algorithms on the EGL-G dataset.

As the problem size grows, solving the problem as a whole becomes much less effective, and divide-and-conquer strategy can be a promising technique in this case. The divide-and-conquer strategy has achieved great success in a range of problems, such as continuous optimization \cite{yang2008large,tan2006distributed,goh2009competitive}, vehicle routing \cite{ostertag2009POPMUSIC,oliveira2016cooperative} and job shop scheduling \cite{zhang2010divide-and-conquer,su2014automatic}.  Also for LSCARP, there have been a variety of divide-and-conquer approaches \cite{mei2013decomposing,mei2014cooperative,shang2016improved,shang2017quantum-Inspired,shang2017memetic} based on decomposing the problem into smaller sub-problems by grouping the routes of the best-so-far solution. Specifically, in these algorithms, the entire search process is divided into cycles. At the beginning of each cycle, the routes of the best-so-far solution are grouped together based on different strategies (e.g. randomly or by clustering methods). They have achieved much better results than the methods without divide-and-conquer.

In 2017, based on two major Chinese cities, i.e. Beijing and Hefei, Tang et al. \cite{tang2017scalable} created two real-world LSCARP datasets, and extended the number of tasks to over 3000. The Beijing and Hefei datasets are much larger than the EGL-G dataset. The existing divide-and-conquer approaches are not effective enough to solve them. Tang et al. \cite{tang2017scalable} proposed a new hierarchical decomposition based on the concept of ``virtual task''. A virtual task is a sequence of tasks. Starting from elementary tasks, the hierarchical decomposition recursively concatenates (virtual) tasks together to form new higher-level virtual tasks. The resultant algorithm, named SAHiD, managed to obtain much better results than other divide-and-conquer approaches (e.g. RDG-MAENS \cite{mei2014cooperative}) within a limited time budget.

In 2017, Kiilerich and W{\o}hlk \cite{kiilerich2017new} generated a very large scale dataset (denoted as the KW set) based on the five countries in Denmark. In the KW dataset, the number of tasks is further enlarged to over 8000. The KW set is the largest dataset by far. W{\o}hlk et al. \cite{wohlk2018fast} proposed a fast heuristic named \emph{Fast-CARP} to solve the LSCARP, in which the whole problem is partitioned into a number of districts and each district is then optimized independently. Fast-CARP can be regarded as a divide-and-conquer approach.

In summary, divide-and-conquer approaches have achieved great success in solving LSCARP effectively. However, the current approaches still have limitations. They focused on the interactions between the tasks in different routes of the best-so-far solution during the decomposition, while the interactions between tasks within each route were neglected. In this paper, we aim to consider the interactions both in different routes and within the same route, and propose the RCO operator for this purpose.

\section{Proposed Route Cutting Off Decomposition}
\label{sec:RCO}

In an adaptive decomposition approach, the entire search process consists of a number of cycles. At the beginning of each cycle, a new decomposition is generated based on the latest information, e.g. typically the best-so-far solution. Specifically, the decomposition tends to assign two tasks into the same subset if (1) the two tasks are close to each other and (2) the two tasks are in the same route of the best-so-far solution. To achieve this, existing works considered to cluster the routes or sub-routes of the best-so-far solution.

When clustering the routes together (e.g. \cite{mei2014cooperative,shang2016improved,shang2017memetic}), the advantage is that the optimal solution under the new decomposition is guaranteed to be no worse than the current best-so-far solution. However, it cannot identify and take advantage of the \emph{patterns} within each route. For example, it is inevitable to have both ``\emph{good links}'' (connecting the tasks that are close to each other) and ``\emph{poor links}'' (connecting the tasks that are distant from each other) in a route, especially in the early stage of the search. Different links are not distinguished when the routes are clustered as a whole.

On the other hand, one can split a route by breaking the poor links. This way, the links between tasks can potentially be treated more properly, although the monotonically non-increasing decomposition cannot be guaranteed. That is, the optimal solution under the new decomposition may be worse than the current best-so-far solution. However, one may identify better decomposition more efficiently. For example, \cite{tang2017scalable} randomly split a route into two sub-routes during the decomposition, and managed to obtain much better results within a limited time budget than the approaches that cluster the whole routes (e.g. \cite{mei2014cooperative}). \cite{xing2010evolutionary} preserved the promising links by counting the total number of appearances of links and breaking those with small appearances.

In this paper, we investigate the \emph{link patterns} within a route more systematically, and propose a \emph{task rank matrix} to represent such patterns. Then, to further improve the effectiveness and efficiency of divide-and-conquer, we propose the RCO decomposition operator based on the task rank matrix.

\subsection{Task Rank Matrix}

First, we define a \emph{link} from one task to another as the shortest path from the former task to the latter task. The direction of the two tasks are discarded. The cost of the link from task $t_1$ to task $t_2$ is defined as:
\begin{gather}
\Delta(t_1, t_2) = \frac{1}{4}\Big(\delta(hv(t_1), hv(t_2)) + \delta(hv(t_1), tv(t_2)) + \delta(tv(t_1), hv(t_2)) + \delta(tv(t_1), tv(t_2))\Big), \label{eq:atd}
\end{gather}
\noindent where $\delta(v_1,v_2)$ indicates the cost of the shortest path from vertices $v_1$ to $v_2$. The quality of a link depends not only on its absolute cost (i.e. how close the two tasks are to each other), but also on the relative cost to other relevant links (i.e. how close they are \emph{in comparison with other alternative tasks}). For example, if a task is isolated and is far away from all the other tasks, then all the links from the task have large costs. However, a link from the isolated task should still be considered as ``good'' if its cost is much smaller than that of the other links from this isolated task.

Based on the above intuition, we define the \emph{task rank matrix} ($\boldsymbol{\Gamma}$) to indicate the quality of the links from each task. Given a set of tasks $T$, the entry $\boldsymbol{\Gamma}_{t_1,t_2}$ represents the rank of the link from task $t_1$ to task $t_2$, which is calculated based on the links from $t_1$ to all the other tasks.

\begin{figure}[!ht]
	\centering
  	\includegraphics[width=0.5\textwidth]{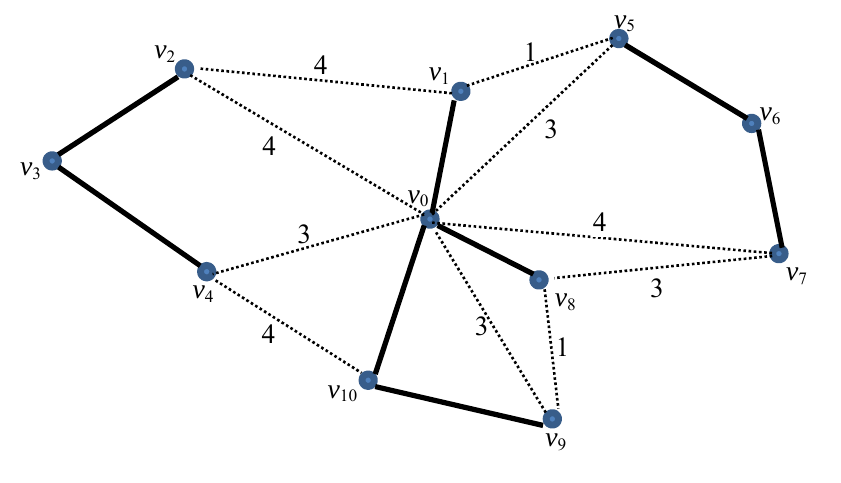}
    \caption{An example to show the calculation of the task rank matrix.}
    \label{fig:example}
\end{figure}

An example is given in Fig. \ref{fig:example} to show how to calculate the task rank matrix. In the figure, $v_0$ is the depot. There are 8 tasks (represented by the solid lines) require to be served, and the shortest paths between the tasks are represented as dashed lines. The service cost and deadheading cost of each task are 1, and the cost of the shortest paths between the tasks are given next to the dashed lines.

For the graph shown in Fig. \ref{fig:example}, we first calculate the matrix of the link costs in Eq. (\ref{eq:linkmatrix}).
\begin{figure*}[!ht]
\footnotesize
\newcounter{TempEqCnt}
\setcounter{TempEqCnt}{\value{equation}}
\begin{equation}
    \mathbf{\Delta} = \bordermatrix{
    ~ & (v_0,v_1) & (v_0,v_8) & (v_0,v_{10}) & (v_2,v_3) & (v_3,v_4) & (v_5,v_6) & (v_6,v_7) & (v_9,v_{10}) \cr
    (v_0,v_1) & - & 4/4 & 4/4 & 18/4 & 16/4 & 8/4 & 12/4 & 8/4 \cr
    (v_0,v_8) & 4/4 & - & 4/4 & 20/4 & 16/4 & 12/4 & 14/4 & 6/4 \cr
    (v_0,v_{10}) & 4/4 & 4/4 & - & 18/4 & 16/4 & 12/4 & 16/4 & 4/4 \cr
    (v_2,v_3) & 18/4 & 20/4 & 18/4 & - & 4/4 & 24/4 & 28/4 & 22/4 \cr
    (v_3,v_4) & 16/4 & 16/4 & 16/4 & 4/4 & - & 24/4 & 28/4 & 20/4 \cr
    (v_5,v_6) & 8/4 & 12/4 & 12/4 & 24/4 & 24/4 & - & 4/4 & 16/4 \cr
    (v_6,v_7) & 12/4 & 14/4 & 16/4 & 28/4 & 28/4 & 4/4 & - & 18/4 \cr
    (v_9,v_{10}) & 8/4 & 6/4 & 4/4 & 22/4 & 20/4 & 16/4 & 18/4 & - \cr
    }. \label{eq:linkmatrix}
\end{equation}
\end{figure*}
\setcounter{equation}{\value{TempEqCnt}}
Then for each row of Eq. (\ref{eq:linkmatrix}), we assign the ranks to the links based on their costs, e.g. the link with the lowest cost is given rank 1. If multiple links have the same cost, then they share the same rank. Eq. (\ref{eq:trm}) shows the task rank matrix obtained from Eq.(\ref{eq:linkmatrix}).
\begin{figure*}[!ht]
\footnotesize
\newcounter{TempEqCnt1}
\setcounter{TempEqCnt1}{\value{equation}}
\setcounter{equation}{10}
\begin{equation}
    \boldsymbol{\Gamma} = \bordermatrix{
    ~ & (v_0,v_1) & (v_0,v_8) & (v_0,v_{10}) & (v_2,v_3) & (v_3,v_4) & (v_5,v_6) & (v_6,v_7) & (v_9,v_{10}) \cr
    (v_0,v_1) & - & 1 & 1 & 7 & 6 & 3 & 5 & 3 \cr
    (v_0,v_8) & 1 & - & 1 & 7 & 6 & 4 & 5 & 3 \cr
    (v_0,v_{10}) & 1 & 1 & - & 7 & 5 & 4 & 5 & 1 \cr
    (v_2,v_3) & 2 & 4 & 2 & - & 1 & 6 & 7 & 5 \cr
    (v_3,v_4) & 2 & 2 & 2 & 1 & - & 6 & 7 & 5 \cr
    (v_5,v_6) & 2 & 3 & 3 & 6 & 6 & - & 1 & 5 \cr
    (v_6,v_7) & 2 & 3 & 4 & 6 & 6 & 1 & - & 5 \cr
    (v_9,v_{10}) & 3 & 2 & 1 & 7 & 6 & 4 & 5 & - \cr
    }. \label{eq:trm}
\end{equation}
\end{figure*}
\setcounter{equation}{\value{TempEqCnt1}}

From the first row of Eq. (\ref{eq:trm}), one can see that the links from $(v_0,v_1)$ to $(v_0,v_8)$ and $(v_0,v_{10})$ are both of rank 1 among all the links from $(v_0,v_1)$, and the link to $(v_2,v_3)$ is of rank 7, since $(v_2,v_3)$ is the farthest task from $(v_0,v_1)$. Note that the task rank matrix is not symmetric. For $(v_0,v_1)$, the task $(v_2,v_3)$ has a very low priority (rank 7), since there are many other closer tasks for $(v_0,v_1)$ to consider. On the contrary, for $(v_2,v_3)$, $(v_0,v_1)$ is a very promising task to go to (rank 2).

\subsection{Route Cutting Off Decomposition Operator}

Based on the task rank matrix, we propose the new RCO operator for decomposition. Briefly speaking, given a best-so-far solution, the RCO operator splits the routes of the solution based on the task rank matrix, to generate a pool of sub-routes. Clustering these sub-routes is expected to obtain more promising decomposition than the existing decomposition methods.

\begin{figure}[!ht]
	\centering
  	\includegraphics[width=0.45\textwidth]{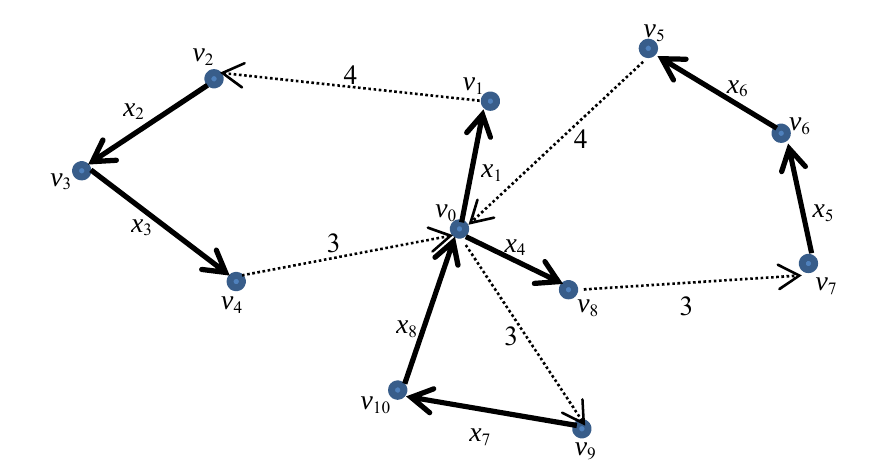}
    \caption{A solution with three routes to the example graph in Fig. \ref{fig:example}.}
    \label{fig:solution}
\end{figure}

The pseudo code of the RCO operator is described in Algorithm \ref{algo:rco}. Given the best-so-far solution $S$, the average task rank $\overline{\gamma}(S)$ of the links in $S$ is first calculated based on the task rank matrix $\boldsymbol{\Gamma}$ (line \ref{line:atr}). For each route $S_k$, we categorize the links into ``good links'' and ``poor links'' (lines \ref{line:gplstart}--\ref{line:gplend}). Here we adopt a simple rule for the categorization. A link is considered to be good if its rank is smaller than $\overline{\gamma}(S)$, and poor otherwise. Then, we randomly cut off a good link with probability $\lambda$, and a poor link with probability $\theta$ (lines \ref{line:cutstart}--\ref{line:cutend}). The corresponding sub-routes are finally inserted into $\Omega$.
\begin{algorithm}[!ht]
\caption{$\mathtt{RCO}(S, \boldsymbol{\Gamma}, \lambda, \theta$)}
\label{algo:rco}
\begin{algorithmic}[1]
\REQUIRE The best-so-far solution $S$, the task rank matrix $\boldsymbol{\Gamma}$, cutting probabilities $\lambda$, $\theta$;
\ENSURE A set of sub-routes $\Omega$;
\STATE Set $\Omega = \emptyset$;
\STATE Calculate the average task rank $\overline{\gamma}(S)$ of $S$ based on $\boldsymbol{\Gamma}$; \label{line:atr}
\FOR{each route $S_k$ of $S$} \label{line:gplstart}
\STATE Set the good links $\mathcal{GL} = \emptyset$, and the poor links $\mathcal{PL} = \emptyset$;
\FOR{each link $\langle S_k[i],S_k[i+1]\rangle$ in $S_k$}
\IF{$\boldsymbol{\Gamma}_{t,t'} < \overline{\gamma}(S)$}
\STATE $\mathcal{GL} = \mathcal{GL} \cup \langle S_k[i],S_k[i+1]\rangle$;
\ELSE
\STATE $\mathcal{PL} = \mathcal{PL} \cup \langle S_k[i],S_k[i+1]\rangle$;
\ENDIF
\STATE Set cut-off good link $gl = null$, cut-off poor link $pl = null$; \label{line:cutstart}
\STATE Randomly generate a number $r_1 \in [0,1]$;
\IF{$r_1 < \lambda$}
\STATE Randomly select the cut-off good link $gl$ from $\mathcal{GL}$;
\ENDIF
\STATE Randomly generate a number $r_2 \in [0,1]$;
\IF{$r_2 < \theta$}
\STATE Randomly select the cut-off poor link $pl$ from $\mathcal{PL}$;
\ENDIF
\STATE Cut off $gl$ and $pl$ in $S_k$ to obtain sub-routes \{$S_{k,1}, \dots\}$; \label{line:cutend}
\STATE $\Omega = \Omega \cup \{S_{k,1}, \dots\}$;
\ENDFOR \label{line:gplend}
\ENDFOR
\STATE {\bf return} $\Omega$;
\end{algorithmic}
\end{algorithm}
Fig. \ref{fig:solution} shows an example solution to the graph shown in Fig. \ref{fig:example}, where the tasks are denoted as $x_1$ to $x_8$. Only the links between the tasks in the same route are considered (e.g. there is no link from $x_3$ to $x_4$). Based on Eq. (\ref{eq:trm}), the ranks of the 5 links are $\boldsymbol{\Gamma}_{x_1, x_2} = 7$, $\boldsymbol{\Gamma}_{x_2, x_3} = 1$, $\boldsymbol{\Gamma}_{x_4, x_5} = 5$, $\boldsymbol{\Gamma}_{x_5, x_6} = 1$, and $\boldsymbol{\Gamma}_{x_7, x_8} = 1$. In this case, the average task rank of the solution is $(7+1+5+1+1)/5 = 3$. $\langle x_2,x_3 \rangle$, $\langle x_5,x_6 \rangle$ and $\langle x_7,x_8 \rangle$ are good links, while $\langle x_1,x_2 \rangle$ and $\langle x_4,x_5 \rangle$ are poor links.
\subsection{Divide-and-Conquer using Route Cutting Off}

Given the best-so-far solution, the proposed RCO operator identifies promising cutting points to split the routes of the solution into promising sub-routes for clustering. In other words, the RCO operator is a generic decomposition operator, and can be embedded into any divide-and-conquer algorithm based on clustering the routes/sub-routes. In this paper, we embed the RCO operator into two state-of-the-art algorithms, i.e. RDG-MAENS \cite{mei2014cooperative} and SAHiD \cite{tang2017scalable} to verify the effectiveness of the RCO operator. The resultant algorithms are named RCO-RDG-MAENS and RCO-SAHiD, respectively.

The pseudo code of RCO-RDG-MAENS is described in Algorithm \ref{algo:rrm}. The algorithm is the same as the original one \cite{mei2014cooperative}, except that instead of clustering the routes directly, the RCO operator is used to split the best-so-far solution before clustering (line \ref{line:rrm-rco}).
\begin{algorithm}[!ht]
\caption{The pseudo code of RCO-RDG-MAENS}
\label{algo:rrm}
\begin{algorithmic}[1]
\REQUIRE The input instance with task set $T$, group number $g$, RCO parameters $\lambda$ and $\theta$;
\ENSURE The best-found feasible solution $S^*$;
\STATE Calculate the task rank matrix $\boldsymbol{\Gamma}$;
\STATE Initialize a population $pop(T)$;
\STATE Evaluate each individual in $pop(T)$;
\STATE Set $S^* = \arg\min_{S \in pop(T)}\{tc(S)\}$;
\FOR {$cycle = 1  \rightarrow max\_cycle$}
\STATE $\Omega = \mathtt{RCO}(S^*, \boldsymbol{\Gamma}, \lambda, \theta$);  \label{line:rrm-rco}
\STATE Cluster the sub-routes in $\Omega$ using the fuzzy $k$-medoid method in RDG \cite{mei2014cooperative} to obtain $\{T_1, T_2, \ldots, T_g\}$;
\FOR {$i = 1 \rightarrow g$}
\STATE $pop(T_i) = \texttt{pop2subpop}(pop(T), T_i)$; \label{line:pop2subpop}
\STATE Evolve $pop(T_i)$ by MAENS \cite{tang2009memetic};
\ENDFOR
\STATE $pop(T) = \texttt{subpop2pop}(pop(T_1), \dots, pop(T_g))$;
\STATE $S' = \arg\min_{S \in pop(T)}\{tc(S)\}$;
\IF {$tc(S') < tc(S^*)$}
\STATE $S^* = S'$;
\ENDIF
\ENDFOR
\STATE {\bf return} $S^*$;
\end{algorithmic}
\end{algorithm}

Likewise, the pseudo code of RCO-SAHiD is described in Algorithm \ref{algo:rsahid}, where the key difference from SAHiD \cite{tang2017scalable} is that the routes of the current solution $S$ are split by RCO (line \ref{line:rsahid-rco}), while they are randomly split in SAHiD.
\begin{algorithm}[!ht]
\caption{The pseudo code of RCO-SAHiD}
\label{algo:rsahid}
\begin{algorithmic}[1]
\REQUIRE The input instance with task set $T$, RCO parameters $\lambda$ and $\theta$
\ENSURE A best feasible solution $S^*$;
\STATE Calculate the task rank matrix $\boldsymbol{\Gamma}$;
\STATE Generate an initial solution $S$ with HDU($T$) \cite{tang2017scalable} operator;
\STATE Improve $S$ with local search procedure;
\STATE Set $S^* = S$;
\WHILE {termination conditions are not satisfied}
\STATE $\Omega = \texttt{RCO}(S, \boldsymbol{\Gamma}, \lambda, \theta)$;  \label{line:rsahid-rco}
\STATE Construct virtual task set $VT$ based on the sub-routes in $\Omega$;
\STATE Call HDU($VT$) to generate a solution $S^{'}$;
\STATE Improve $S^{'}$ with local search procedure;
\IF {$S^{'}$ is acceptable}
\STATE $S = S^{'}$;
\IF {$tc(S^{'}) < tc(S^*)$}
\STATE $S^* = S^{'}$;
\ENDIF
\ENDIF
\ENDWHILE
\STATE {\bf return} $S^*$;
\end{algorithmic}
\end{algorithm}

\section{Experimental Studies}
\label{sec:exp}
To evaluate the effectiveness of the proposed RCO operator for LSCARP, we conduct experiments to compare RCO-RDG-MAENS and RCO-SAHiD with their original counterparts as well as other state-of-the-art algorithms on a range of LSCARP instances. In addition to RDG-MAENS and SAHiD, we also compare with VNS \cite{polacek2008variable}, TSA1 \cite{brandao2008deterministic}, ILS-RVND \cite{martinelli2011improved}, IRDG-MAENS \cite{shang2016improved}, QICA-CARP \cite{shang2017quantum-Inspired}, ESMEANS \cite{shang2017memetic}, Fast-CARP \cite{wohlk2018fast}, PS  \cite{zbib2017vriants} and UHGS \cite{vidal2017node}.

\subsection{Datasets}

Since our work focuses on LSCARP, we select the four existing LSCARP datasets, i.e. the EGL-G \cite{brandao2008deterministic}, Hefei \cite{tang2017scalable}, Beijing \cite{tang2017scalable} and KW  \cite{kiilerich2017new} datasets. The EGL-G dataset consists of 10 instances, which are derived from a real-world road network of Lancashire, UK, with 255 nodes and 375 edges. The dataset contains two groups G1 and G2, each with 5 instances. The instances belonging to the same group have the same task set, and different vehicle capacities. The Hefei dataset contains 10 instances based on a road network in Hefei, China, with 850 nodes and 1212 edges. The instances have the same vehicle capacity, but vary in their task sets. Similarly, the Beijing dataset consists of 10 instances sharing the same road network in Beijing, China (2820 nodes and 3584 edges) and vehicle capacity. The KW dataset consists of 264 CARP benchmark instances generated from 88 graphs by varying the vehicle capacity. In the KW dataset, the largest instance contains 11640 nodes, 12675 edges, and 8581 required edges. Overall, EGL-G is the smallest LSCARP dataset. Hefei and Beijing are much larger than EGL-G, and KW is the largest dataset.

\subsection{Experiment Design}

Since different datasets have different available results, we designed three experimental comparisons as follows.
\begin{itemize}
    \item {\bf Experiment 1}: compare \emph{RCO-RDG-MAENS} with RDG-MAENS \cite{mei2014cooperative}, ILS-RVND \cite{martinelli2011improved}, IRDG-MAENS \cite{shang2016improved}, QICA-CARP \cite{shang2017quantum-Inspired} and ESMAENS \cite{shang2017memetic} on the EGL-G dataset.
    \item {\bf Experiment 2}: compare \emph{RCO-SAHiD} with SAHiD \cite{tang2017scalable},  UHGS \cite{vidal2017node}, RDG-MAENS \cite{mei2014cooperative}, VNS \cite{polacek2008variable} and TSA1 \cite{brandao2008deterministic} on the Hefei and Beijing datasets.
    \item {\bf Experiment 3}: compare \emph{RCO-SAHiD} with SAHiD \cite{tang2017scalable}, UHGS \cite{vidal2017node}, Fast-CARP \cite{wohlk2018fast} and PS \cite{zbib2017vriants} on some large KW instances (task number ranging 7831 to 8581).
\end{itemize}

In each experiment, we tried our best to compare with all the algorithms whose results are available in literature for the corresponding datasets. For each instance, each algorithm was run multiple times independently, and the Wilcoxon rank sum test \cite{wilcoxon1945individual} was conducted to test the results statistically.

To make fair comparisons, we set the algorithm parameters consistent with the settings in the literature. Specifically, in Experiment 1, following the parameter settings in \cite{mei2014cooperative}, in RCO-RDG-MAENS, we set the population size to $30$, offspring population size to $180$, maximum generation number to $500$, number of cycles to $50$, and probability of local search to $0.2$. The RDG operator has two parameters: the number of groups $g$ and degree of fuziness $\alpha$ (used by the fuzzy $k$-medoid clustering). We set $g = 2$ and $\alpha = 5$ for both RCO-RDG-MAENS and RDG-MAENS, as they showed the best performance \cite{mei2014cooperative}.

In Experiment 2, following the parameter settings in \cite{tang2017scalable}, in RCO-SAHiD, the scale parameter in HD is $0.1$, the threshold for accepting a worse solution is $110\%$, and the maximum number of idle iterations for accepting an ascending move is $10000$. Note that in Experiment 2, the stopping criterion of RCO-SAHiD and UHGS is the runtime i.e. after 30 minutes. The runtime depends on a variety of factors such as CPU frequency, RAM, operating system, coding language and compiler. In our experiments, we implemented RCO-SAHiD based on the original SAHiD source code to make sure they share the same programming language and compiler. To improve fairness, a common approach used by previous studies (e.g.\cite{mei2011memetic,zhang2017memetic,tang2009memetic,mei2009global,martinelli2011improved,mei2014cooperative}) is to scale the runtime based on the CPU frequency. In this paper, we adopt the same scaling approach. RCO-SAHiD and UHGS were run on Intel(R) Xeon(R) E5-2650 v2 with 2.6 GHz, 64GBs RAM, and the other compared algorithms were run on Intel Core i7-4790 with 3.6 GHz. Therefore, the maximum runtime for RCO-SAHiD and UHGS was set to $3.6/2.6 \times 30 \times 60 = 2492$ seconds.

{
In Experiment 3, the parameters of RCO-SAHiD are the same as in Experiment 2. The runtime follows the configuration of the original literature of Fast-CARP \cite{wohlk2018fast}, which is one minute per 1000 nodes. Fast-CARP was run on an Intel Xeon CPU with 3.5 GHz, while RCO-SAHiD, SAHiD and UHGS were run on Intel(R) Xeon(R) E5-2650 v2 with 2.6 GHz, 64GBs RAM. The runtime of these three algorithms were scaled to $3.5/2.6 \times 60 = 81$ seconds per 1000 nodes.

\subsection{Parameter Sensitivity Analysis}

The RCO operator has two important parameters, namely the probability of cutting good links $\lambda$ and the probability of cutting poor links $\theta$. Intuitively, it is more promising to cut poor links than good links. On the other hand, cutting good links can increase the exploration capability of the algorithm, and help the search jump out of the current local optimum. Based on the above consideration, we should set $\lambda$ to a small value, and $\theta$ to a relatively large value. To analyze the sensitivity of $\lambda$ and $\theta$, we conducted some pilot experiments by running RCO-RDG-MAENS with $\lambda = 0.05$ and $0.1$, and $\theta = 0.1$, $0.2$ and $0.3$ (six combinations in total) on the EGL-G dataset. Each algorithm was run 30 times independently on each instance.

Fig. \ref{fig:RCO_tc} and Fig. \ref{fig:RCO_time} show the average total cost and computational time of RCO-RDG-MEANS with different values of $\lambda$ and $\theta$ over the 10 EGL-G instances. From the figures, one can see that for all the instances, there is no significant difference among the different $\lambda$ and $\theta$ values in terms of the average total cost. However, the $\lambda$ and $\theta$ values have some impact on the computational time. The algorithm with $\lambda = 0.1$ and $\theta = 0.3$ usually had the longest computational time. All the other five versions had similar computation time. Overall, the algorithm with $(\lambda,\theta) = (0.05,0.2)$ seems to have a short computational time over all the instances.
\begin{figure*}[!ht]
	\centering
  	\includegraphics[width=0.75\textwidth]{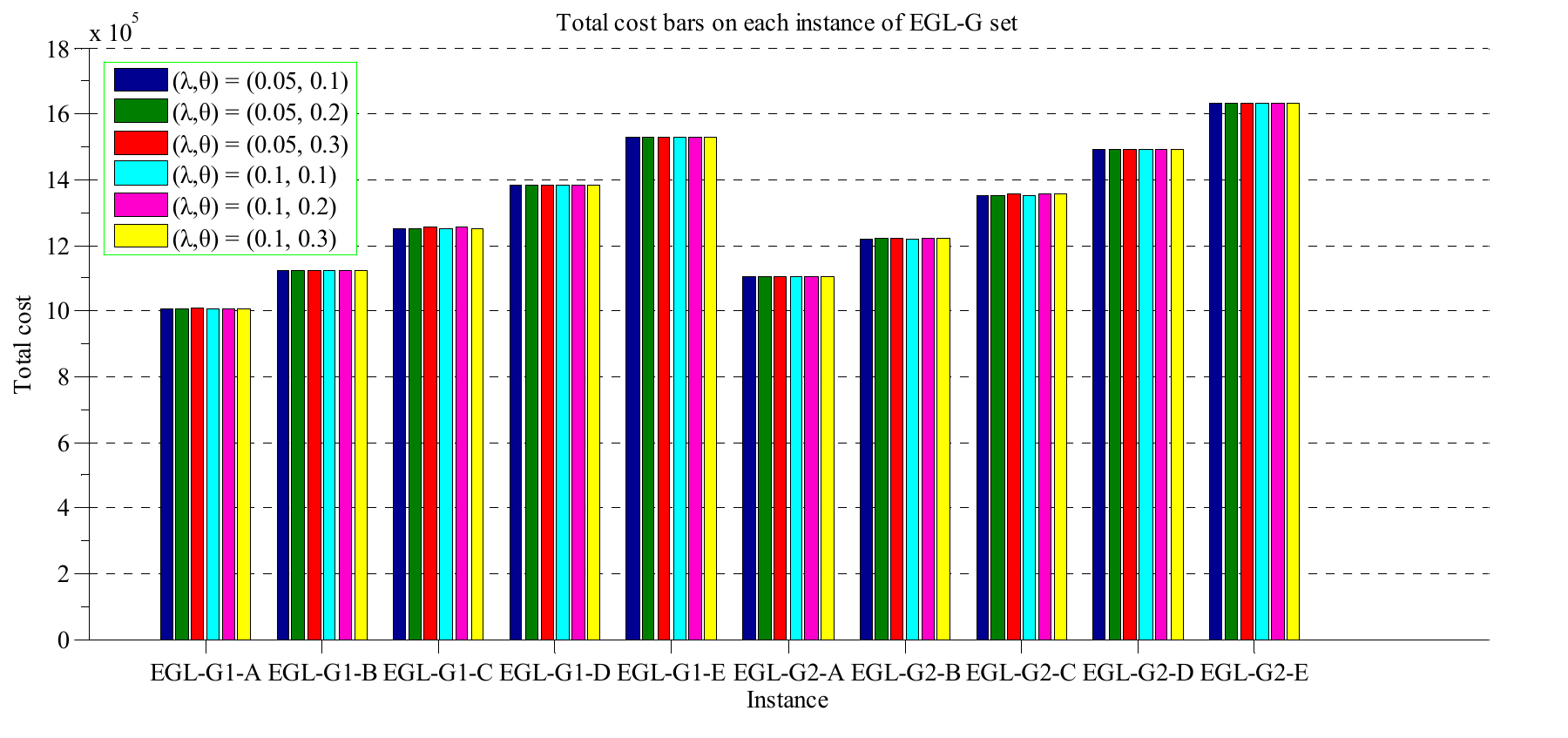}
    \caption{The average total cost over the 30 independent runs of RCO-RDG-MEANS on each EGL-G instance.}
    \label{fig:RCO_tc}
\end{figure*}
\begin{figure*}[!ht]
	\centering
  	\includegraphics[width=0.75\textwidth]{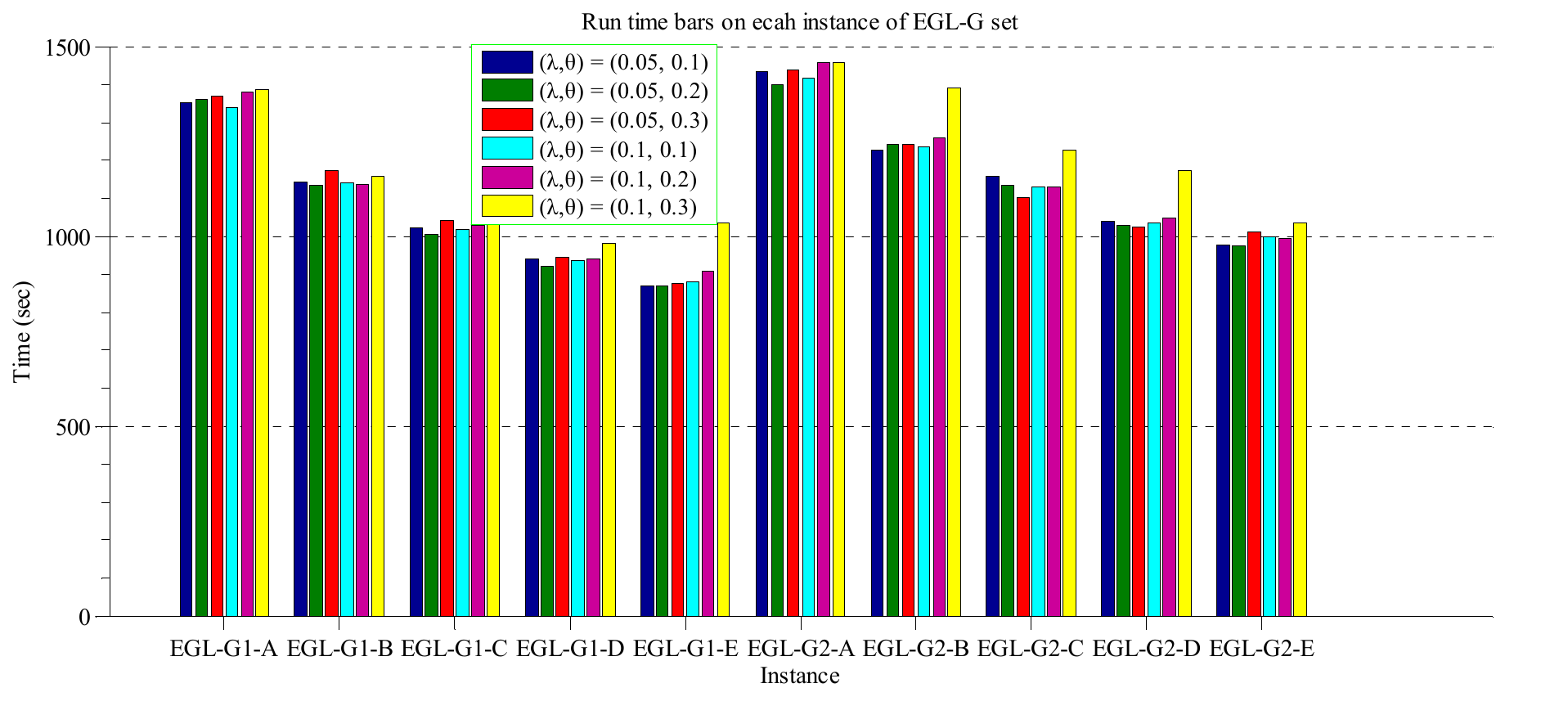}
    \caption{The average computational time over the 30 independent runs of RCO-RDG-MEANS on each EGL-G instance.}
    \label{fig:RCO_time}
\end{figure*}


In summary, the tested $\lambda$ and $\theta$ values have little impact on the performance, and affect the computational time only slightly. This is a good sign, as it means that the effectiveness and efficiency of the algorithm are not sensitive to the $\lambda$ and $\theta$ values within a considerable range.

Fig. \ref{fig:RCO_time} shows that $(\lambda,\theta) = (0.05,0.2)$ achieved the overall lowest computation time. Therefore, in the subsequent experiments, we select $(\lambda,\theta) = (0.05,0.2)$ for the RCO operator in both RCO-RDG-MAENS and RCO-SAHiD.

\subsection{Results on Experiment 1}

In Experiment 1, RCO-RDG-MAENS is compared with RDG-MAENS \cite{mei2014cooperative}, ILS-RVND \cite{martinelli2011improved}, IRDG-MAENS \cite{shang2016improved}, QICA-CARP \cite{shang2017quantum-Inspired}, and ESMEANS \cite{shang2017memetic}. For RDG-MAENS, we downloaded the code from online \footnote{The C code of RDG-MAENS is available from \url{http://homepages.ecs.vuw.ac.nz/~yimei/codes/RDG-MAENS.zip}} and reran it for 30 times independently on each instance. However, for all the other compared algorithms, no code is available for rerunning the experiments. Therefore, we directly copied the results of these algorithms from their original literature. Note that we configured RCO-RDG-MAENS in exactly the same way as RDG-MAENS. Therefore, we can guarantee a fair comparison with RDG-MAENS and other algorithms (as they compared with RDG-MAENS under the same configuration).

Table \ref{table:rdg-eglg-avg} shows the average performance of the compared algorithms on the 10 EGL-G instances. In the table, the columns ``$|V|$'', ``$|E|$'' and ``$|T|$'' stand for the number of vertices, edges and tasks, respectively. $\zeta$ indicates the minimum number of vehicles required to serve all the routes, which can be computed as $\zeta = \left\lceil\frac{\sum_{t \in T}{d(t)}}{Q}\right\rceil$, where $d(t)$ is the demand of task $t$ and $Q$ refers to the capacity of the vehicles. In general, with the same problem size, a larger $\zeta$ value implies a more complex problem instance. For each algorithm, the columns ``Mean'' and ``Std'' are the mean and standard deviation of the total costs obtained by 30 independent runs. Note that there is no ``Std'' column for ILS-RVND and IRDG-MAENS, since only the mean value was reported in their original literature.

For each instance, the minimal mean total cost among all the algorithms is marked with ``$\dagger$''. In addition, we conduct statistical test between RCO-RDG-MAENS and each compared algorithm using Wilcoxon rank sum test under the significance level of $0.05$. If an algorithm is significantly worse (better) than RCO-RDG-MAENS, then it is marked with underline (in bold). In addition, the last row ``W-D-L'' stands for the number of instances on which RCO-RDG-MAENS performed significantly better than (``W''), statistically comparable with (``D''), and significantly worse than (``L'') the corresponding algorithm. For example, ``1-9-0'' under RDG-MAENS indicates that RCO-RDG-MAENS performed significantly better than RDG-MAENS on 1 instance, and statistically comparable with it on the remaining 9 instances.

From Table \ref{table:rdg-eglg-avg}, RCO-RDG-MAENS performed statistically comparable with RDG-MAENS on 9 out of 10 EGL-G instances, and significantly outperformed it on G2-B.
In comparison with the other algorithms, RCO-RDG-MAENS performed much better. It significantly outperformed ILS-RVND and QICA-CARP on all the instances, and IRDG-MAENS and ESMAENS on 7 and 8 instances, respectively. RCO-RDG-MAENS never performed significantly worse than any compared algorithm on any instance.

\begin{table*}[!ht]
\addtolength{\tabcolsep}{-2.8pt}
\footnotesize
\caption{The average performance over 30 independent runs of the compared algorithms on the EGL-G dataset. For each instance, the minimal mean total cost is marked with ``$\dagger$''. Under Wilcoxon rank sum test with significance level of $0.05$, an algorithm is marked with underline (in bold) if it is significantly worse (better) than RCO-RDG-MAENS.}
  \label{table:rdg-eglg-avg}
\begin{center}
    \begin{tabular}{@{}ccccccccccccccc@{}}
    \toprule
Name & $|V|$ & $|E|$ & $|T|$ & $\zeta$ & ILS-RVND& IRDG-MAENS & \multicolumn{2}{c}{QICA-CARP} & \multicolumn{2}{c}{ESMAENS} & \multicolumn{2}{c}{RDG-MAENS} & \multicolumn{2}{c}{RCO-RDG-MAENS} \\
\cmidrule(lr){6-6} \cmidrule(lr){7-7} \cmidrule(lr){8-9} \cmidrule(lr){10-11}\cmidrule(lr){12-13} \cmidrule(lr){14-15}
    &       &       &       &       & Mean & Mean & Mean & Std & Mean & Std & Mean & Std & Mean & Std \\
    \midrule
    G1-A &        255 &        375 &        347 &         20 & \underline{1010937.4} & { {\underline{1007977.1}}} & { {\underline{1008151.8}}} &     4441.1 & {{\underline{1007807.0}}} &       4462.0 &{ 1007368.0} &4311.2& 1005870.4$^\dagger$  &       3827.0 \\

      G1-B &        255 &        375 &        347 &         25 & \underline{1137141.5} & { {\underline{1125763.6}}} & { {\underline{1125874.0}}} &     5802.8 & { {\underline{1125649.7}}} &     5214.6 & {1123369.1} &5528.0& 1121529.2$^\dagger$  &     4437.7 \\

      G1-C &        255 &        375 &        347 &         30 & {\underline{1266576.8}} & {\underline{1255674.1}} & { {\underline{1252912.8}}} &     5242.1 & {\underline{1254856.3}} &     6233.1 & 1251028.7 &4268.1&1250070.5$^\dagger$  &     4048.2 \\

      G1-D &        255 &        375 &        347 &         35 & {\underline{1406929.0}}& { {\underline{1388277.5}}} & { {\underline{1387461.7}}} &     6012.5 & { {\underline{1385882.0}}} &     4112.6 & 1384901.5 &6131.4& 1383354.8$^\dagger$  &     4390.6 \\

      G1-E &        255 &        375 &        347 &         40 & {\underline{1554220.2}} & { 1528397.0} & { {\underline{1529252.2}}} &     6101.5 & { {\underline{1530893.7}}} &     7361.4 & 1527631.0 &5641.2& 1526502.9$^\dagger$  &     5895.1 \\

      G2-A &        255 &        375 &        375 &         22 &  {\underline{1118363.0}} &  1108959.5 & {\underline{1109462.4}} &     5923.1 & { 1107939.3} &     3282.9 & 1106081.9$^\dagger$  &5144.1 &1106843.3  &     4586.6 \\

      G2-B &        255 &        375 &        375 &         27 & {{\underline{1233720.5}}} & {{\underline{1223541.5}}} & {{\underline{1222531.7}}} &     4843.3 & {{\underline{1223247.4}}} &     5608.4 & {\underline{1223705.7}} &5802.2& 1220453.5$^\dagger$  &     5086.5 \\

      G2-C &        255 &        375 &        375 &         32 & {\underline{1374479.7}} & { 1353653.7} & {\underline{1355637.0}} &     5344.8 & {\underline{1355667.3}} &     5589.5 & 1353819.1 &5169.6& 1352801.7$^\dagger$  &     4289.5 \\

      G2-D &        255 &        375 &        375 &         37 & {\underline{1515119.3}} & {\underline{1495822.2}} & { {\underline{1492428.0}}} &       3696.0 &  { 1492155.9} &     6385.7 & 1492745.4 &7146.8&1490704.2$^\dagger$  &     5973.5 \\

      G2-E &        255 &        375 &        375 &         42 & {\underline{1658378.1}} & { {\underline{1636473.4}}} & {{\underline{1636746.5}}} &     5764.3 & { {\underline{1635161.3}}} &       5737.0 & 1633191.9 &5704.8& 1631377.8$^\dagger$  &     6041.6 \\[5pt]

    W-D-L &       &       &       &       & 10-0-0     & 7-3-0 & 10-0-0 &       & 8-2-0 &&1-9-0 &       &       &  \\

\bottomrule
\end{tabular}
\end{center}
\end{table*}

Table \ref{table:rdg-eglg-best} shows the best total cost obtained from the 30 independent runs of the compared algorithms on each instance.\footnote{{The full detail of the best solutions can be found from \url{https://meiyi1986.github.io/files/data/carp/results.zip}.
}} The minimal total cost is marked with ``$\dagger$'' and result is marked with undeline (in bold) if it is larger (smaller) than that of RCO-RDG-MAENS. In the last two rows, the ``Mean'' row stands for the mean values of the best total costs obtained by each compared algorithm over all the instances, and  the last row ``G-E-S'' stands for the number of instances on which results obtained by RCO-RDG-MAENS was greater than (``G''), equal to (``E'), and smaller than (``S'') the corresponding algorithm. Note that the table does not include IRDG-MAENS \cite{shang2016improved}, since the best performance of IRDG-MAENS was not reported in the original literature.

\begin{table*}[!ht]
\footnotesize
  \caption{The best total cost of the compared algorithms on EGL-G dataset. For each instance, the minimal total cost is marked with ``$\dagger$''. An algorithm is marked with underline (in bold) if its total cost is greater (smaller) than RCO-RDG-MAENS.}
  \label{table:rdg-eglg-best}
\begin{center}
\begin{tabular}{cccccc}
 \toprule

      Name &  ILS-RVND &  QICA-CARP &    ESMAENS & RDG- MAENS & RCO-RDG-MAENS \\

\midrule
      G1-A & \underline{1002264} & \underline{999151}  & \underline{998682}   & \bf{998405}$^\dagger$  & 998763 \\

      G1-B & \underline{1126509}   & 1118030$^\dagger$ & \underline{1118092} & 1118030$^\dagger$ & 1118030$^\dagger$ \\

      G1-C & \underline{1260193} & \underline{1245398}    & \underline{1246350} & \bf{1242897}$^\dagger$ &  1243096 \\

      G1-D & \underline{1397656} & \underline{1376795}  & \underline{1377291} & \underline{1375583}   & 1375319$^\dagger$ \\

      G1-E & \underline{1541853} & \underline{1518055}   & \underline{1516089}  & \underline{1518694}  & 1513589$^\dagger$ \\

      G2-A & \underline{1111127} & \underline{1100447} & \underline{1100134}  & \underline{1097581}   & 1097291$^\dagger$ \\

      G2-B & \underline{1223737}  & \underline{1213004} & \underline{1212564} & \underline{1211805}   & 1211789$^\dagger$ \\

      G2-C & \underline{1366629}  & \underline{1344221}    & \bf{1343044}$^\dagger$& \underline{1344228}    & 1344353 \\

      G2-D & \underline{1506024 } & \underline{1482861} & \bf{1478162}$^\dagger$ & \underline{1482216}  & 1482345 \\

      G2-E & \underline{1650657}  & \underline{1625984}    & \underline{1622275}  & \underline{1622927} & 1621354$^\dagger$ \\[5pt]

      Mean & 1318665.0 &  1302394.6 &  1301268.3 & 1301236.6 & 1300592.9 \\

         G-E-S &10-0-0 &          9-1-0 &          8-0-2&       7-1-2 &    \\
\bottomrule
\end{tabular}
\end{center}
\end{table*}

From Table \ref{table:rdg-eglg-best}, one can see that RCO-RDG-MAENS reached the minimal total cost on 6 out of 10 instances, which is much more than that of the other algorithms (i.e., 3 instances for RDG-MAENS, 0 for ILS-RVND, 1 for QICA-CARP and 2 for ESMAENS). In terms of the mean of the best total cost over the 10 instances, RCO-RDG-MAENS performed much better than the other compared algorithms (e.g. $1300592.9$ versus $1301236.6$ in comparison with RDG-MAENS).

To make more intuitive comparisons between RCO-RDG-MAENS and RDG-MAENS, two representative instances (i.e., G1-A and G2-E) are selected from EGL-G, and the convergence curves of RCO-RDG-MAENS and RDG-MAENS on them  are shown in Fig. \ref{fig:two_figs_EGL}. In the figure, the $x$-axis is the computational time, and the $y$-axis is the mean total cost of the best-so-far solutions obtained by the two algorithms. From the figure, one can see that the convergence curves of the two algorithms are very close to each other. RCO-RDG-MAENS tend to converge slightly slower than RDG-MAENS in the early stage of the search, and then catch up and achieve better results than RDG-MAENS in the later stage (e.g. the two curves crossed each other after around 500 seconds for both instances).
\begin{figure*}[!ht]
	\centering
  	\includegraphics[width=0.45\textwidth]{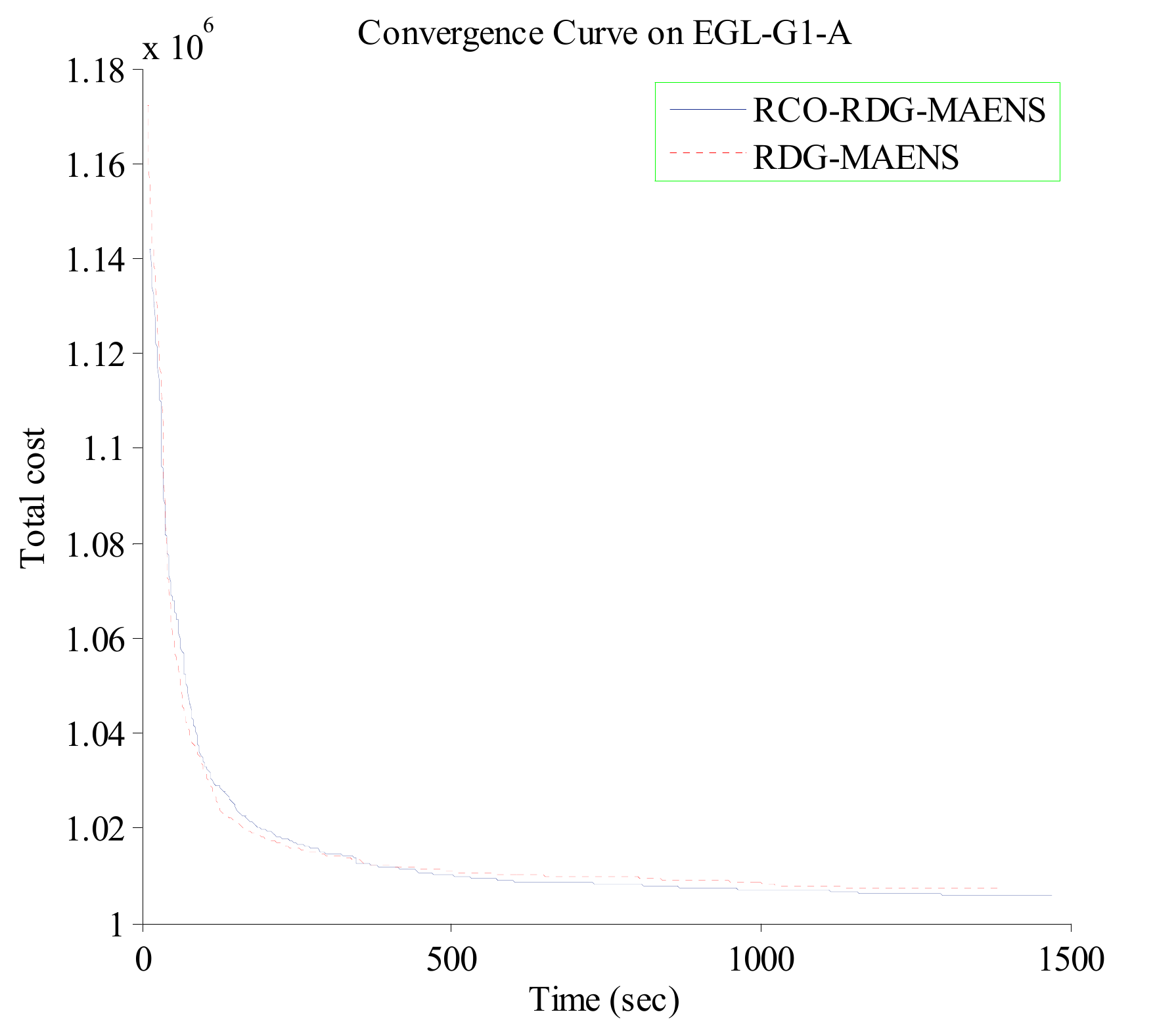}
    \includegraphics[width=0.45\textwidth]{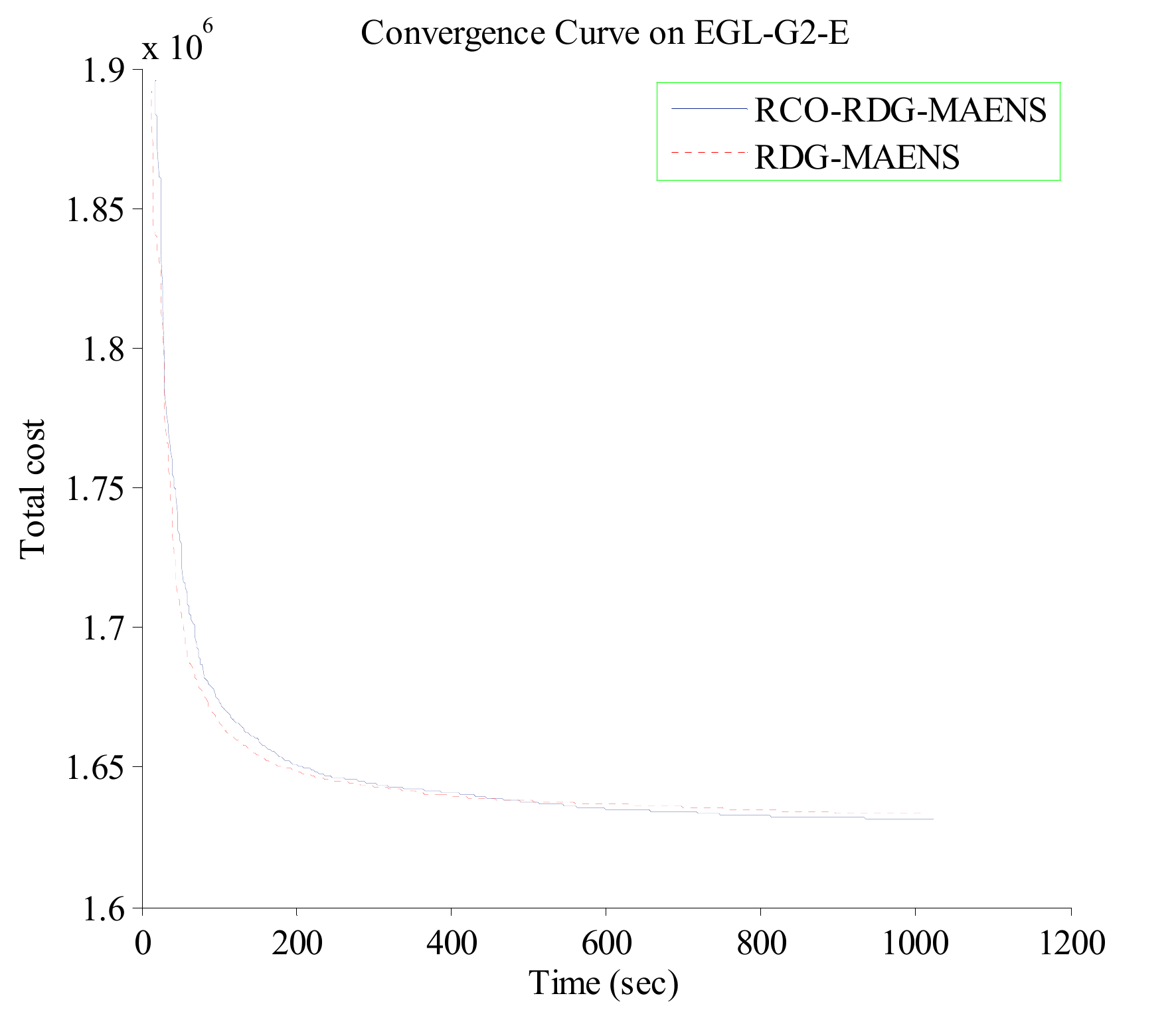}
    \caption{Convergence curves of RCO-RDG-MAENS and RDG-MAENS on instances EGL-G1-A and EGL-G2-E.}
    \label{fig:two_figs_EGL}
\end{figure*}

In summary, the comparison between RCO-RDG-MAENS and other algorithms including RDG-MAENS shows that the RCO operator can improve the performance of RDG-MAENS on the EGL-G instances. In terms of average performance, RCO-RDG-MAENS significantly outperformend RDG-MAENS on 1 instance, and was never beaten by RDG-MAENS. In terms of best performance, RCO-RDG-MAENS managed to outperform RDG-MAENS (and other compared algorithms) on most EGL-G instances.

The advantage of the RCO operator looks marginal in Experiment 1. This is partially because RDG-MAENS already has a high decomposition accuracy for medium-sized instances such as the EGL-G instances \cite{tang2017scalable}, and there is not much space for improvement by the RCO operator. However, the Hefei, Beijing and KW datasets are much larger and more complex. Thus, we expect the RCO operator to show more advantage in Experiments 2 and 3.

\subsection{Results on Experiment 2}

In Experiment 2, RCO-SAHiD is compared with SAHiD \cite{tang2017scalable},  UHGS \cite{vidal2017node}, RDG-MAENS \cite{mei2014cooperative}, VNS \cite{polacek2008variable} and TSA1 \cite{brandao2008deterministic} on the Hefei and Beijing datasets, which are much larger than the EGL-G dataset. Following the same practice in \cite{tang2017scalable}, both RCO-SAHiD and SAHiD were ran 25 times independently. The maximum runtime of the compared algorithms is set to $2492$ seconds after the scaling.

Note that the configuration of Experiment 2 is very different from Experiment 1. The instances in Experiment 2 is much larger than the instances in Experiment 1, and the time budget is much tighter. Therefore, the search efficiency of the algorithm within a very limited time budget becomes much more important.

Table \ref{table:sahid-hefei-ave} shows the average performance of the compared algorithm on the 10 Hefei instances. For each instance, the minimal mean total cost is marked with ``$\dagger$''. Under Wilcoxon rank sum test with significance level of $0.05$, if an algorithm is significantly worse (better) than RCO-SAHiD, then its result is marked with underline (in bold).

From Table \ref{table:sahid-hefei-ave}, one can see that RCO-SAHiD significantly outperforms all the other compared algorithms except UHGS with respect to the average performance. In particular, RCO-SAHiD significantly outperformed SAHiD on 9 out of the 10 instances, with much smaller mean and standard deviation. This indicates that embedding RCO into SAHiD can greatly improve its effectiveness and stability.


\begin{table*}[!t]\addtolength{\tabcolsep}{-2.2 pt}
\footnotesize
  \caption{The average performance over 25 independent runs of the compared algorithms on the Hefei dataset. For each instance, the minimal mean total cost is marked with ``$\dagger$''. Under Wilcoxon rank sum test with significance level of $0.05$, an algorithm is marked with underline (in bold) if it is significantly worse (better) than RCO-SAHiD.}
  \label{table:sahid-hefei-ave}
\begin{center}
\begin{tabular}{@{}ccccccccccccccccc@{}}
\toprule

Name & $|V|$ & $|E|$ & $|T|$ & $\zeta$  & \multicolumn{ 2}{c}{RDG-MAENS} & \multicolumn{ 2}{c}{VNS} & \multicolumn{ 2}{c}{TSA1} & \multicolumn{ 2}{c}{UHGS}& \multicolumn{ 2}{c}{SAHiD} &\multicolumn{ 2}{c}{RCO-SAHiD} \\
\cmidrule(lr){6-7} \cmidrule(lr){8-9} \cmidrule(lr){10-11} \cmidrule(lr){12-13}\cmidrule(lr){14-15}\cmidrule(lr){16-17}

 & &  &  &  &    Mean &        Std &    Mean &        Std &    Mean &Std &    Mean &        Std &    Mean &        Std &    Mean &        Std \\
\midrule
  {Hefei}-1 &        850 &       1212 &        121 &          7 &    247341  &       2293 & { {\underline{247819}}} &       2745 & {\underline{252615}} &       1591 &    {\bf{245596}$^\dagger$} &       0 &   {\underline{251024}} &       1820 & 247351 &      536 \\

  {Hefei}-2 &        850 &       1212 &        242 &         14 & { {\underline{441539}}} &       4142 & {\underline{449979}} &       5375 & {\underline{456228}} &       5539 & {\bf{433807}$^\dagger$} &       99&   {\underline{445376}} &       2476 &437631   &     1208 \\

  {Hefei}-3 &        850 &       1212 &        364 &         19 & { {\underline{589152}}} &       2697 & {\underline{595263}} &       3108 & {\underline{637201}} &       8003 & {\bf{573737}$^\dagger$} &       955 &   {\underline{590969}} &       2305 &586795   &     1241 \\

  {Hefei}-4 &        850 &       1212 &        485 &         28 & {\underline{761351}} &       4362 & {\underline{774323}} &       6394 & {\underline{791790}} &       5481 & {\bf{740404}$^\dagger$} &       1577 &   {\underline{759402}} &       2495 &753859   &     1898 \\

  {Hefei}-5 &        850 &       1212 &        606 &         35 & {\underline{991813}} &       5755 & {\underline{994794}} &       6109 & {\underline{1042701}} &      11496 & {\bf{946574}$^\dagger$} &      1741&   {\underline{976276}} &       4742 &967045   &     2766 \\

  {Hefei}-6 &        850 &       1212 &        727 &         42 & {\underline{1132063}} &       8966 & {\underline{1128667}} &       9404 & {\underline{1162641}} &      13806 & {\bf{1072864}$^\dagger$} &       3024&   {\underline{1106735}} &       5318 &1098915   &     3964 \\

  {Hefei}-7 &        850 &       1212 &        848 &         49 & {\underline{1361125}} &      14356 & {\underline{1337353}} &       6745 & {\underline{1353502}} &       6235 & {\bf{1272880}$^\dagger$} &       3920&   {\underline{1309474}} &       4792 & 1305057   &     3798 \\

  {Hefei}-8 &        850 &       1212 &        970 &         56 & {\underline{1550509}} &      13695 & {\underline{1517151}} &      12477 & {\underline{1537169}} &       6709 & {\bf{1436048}$^\dagger$} &       4838&   {\underline{1483694}} &       4857 &1478098   &     4466 \\

  {Hefei}-9 &        850 &       1212 &       1091 &         63 & {\underline{1749079}} &      18872 & {\underline{1694957}} &      10164 & {\underline{1716256}} &       9236 & {\bf{1605554}$^\dagger$} &       5151&   {\underline{1659700}} &       6103 & 1656147   &     4493 \\

 {Hefei}-10 &        850 &       1212 &       1212 &         69 &   {\underline{1923264}} &      31697 & {\underline{1852622}} &      10183 & {\underline{1901167}} &      12679 & {\bf{1754889}$^\dagger$} &       4306&   1808860  &       7836 & 1810301 &     6003 \\[5pt]
     W-D-L &            &            &            &             & {9-1-0} &            &     10-0-0 &            & { 10-0-0} &&   { 0-0-10} &         &{ 9-1-0} &         &            &            \\
\bottomrule
\end{tabular}
\end{center}
\end{table*}

Table \ref{table:sahid-hefei-best} shows the best total cost obtained by the compared algorithms over the 25 independent runs on the Hefei dataset. The table shows consistent patterns in terms of the best performance with the average performance. UHGS performed the best on all the 10 Hefei instances. It is followed by RCO-SAHiD, which is much better than all the other algorithms.



\begin{table*}[!t]
\footnotesize
  \caption{The best total cost of the compared algorithms on the Hefei dataset. For each instance, the minimal total cost is marked with ``$\dagger$''. An algorithm is marked with underline (in bold) if its total cost is greater (smaller) than RCO-SAHiD.}
  \label{table:sahid-hefei-best}
\begin{center}
\begin{tabular}{ccccccc}
\toprule
      Name &      RDG-MAENS &        VNS &       TSA1 & UHGS &SAHiD & RCO-SAHiD \\
\midrule
 {Hefei}-1 &     \underline{ 246221} &    \bf{245596}$^\dagger$  &     \underline{250155} &     \bf{245596}$^\dagger$& \underline{248048} & 246571 \\

  {Hefei}-2 &    \bf {436020}  & \underline{ 436637} &     \underline{447853} &    \bf{433648}$^\dagger$&\underline{441574} & 436031 \\

  {Hefei}-3 &    \underline{ 583050} &    \underline{588682}  &     \underline{623795} &  \bf{572545}$^\dagger$&\underline{586880} & 582839  \\

  {Hefei}-4 &     \underline{754855} &    \underline{763256}  &     \underline{774182} &    \bf{737730}$^\dagger$&\underline{754015} &750687  \\

  {Hefei}-5 &     \underline{980153} &     \underline{984121} &    \underline{1019224} &   \bf{941278}$^\dagger$&\underline{964772} &{961376}   \\

  {Hefei}-6 &    \underline{1119584} &    \underline{1110030} &    \underline{1134041} &  \bf{1068035}$^\dagger$&\underline{1095530} & 1092667  \\

  {Hefei}-7 &    \underline{1329745} &    \underline{1322290} &    \underline{1339160} &  \bf{1266931}$^\dagger$&\underline{1299430} & 1299360  \\

  {Hefei}-8 &    \underline{1526453} &   \underline{1492790}  &    \underline{1521857} &  \bf{1427531}$^\dagger$&\underline{1474390} & 1469819  \\

  {Hefei}-9 &    \underline{1705381} &   \underline{1675790}  &  \underline{1696706}   &   \bf{1598203}$^\dagger$&\underline{1648840} &1645841  \\

 {Hefei}-10 &   \underline{1837767} &    \underline{1834860} &   \underline{1873504}  &  \bf{1748829}$^\dagger$&\bf{1793890}  &  1799158 \\[5pt]

      Mean &  1051922.9 &  1045405.2 &  1068047.7 & 1004032.6 & 1030736.9 & 1028434.9 \\

        G-E-S &        9-0-1 &          9-0-1 &          10-0-0 & 0-0-10 &         9-0-1 & \\

\bottomrule
\end{tabular}
\end{center}
\end{table*}

Tables \ref{table:sahid-beijing-ave} and \ref{table:sahid-beijing-best} show the average and best performance of the compared algorithms on the Beijing dataset. From the tables, we can observe consistent patterns with those on the Hefei dataset. UHGS performed the best on all the Beijing instances in terms of both average and best performance. RCO-SAHiD was the second best algorithm, showing significantly better performance than all the other compared algorithms (including SAHiD) on all the Beijing instances.


\begin{table*}[!t]\addtolength{\tabcolsep}{-2.9 pt}
\footnotesize
\caption{The average performance over 25 independent runs of the compared algorithms on the Beijing dataset. For each instance, the minimal mean total cost is marked with ``$\dagger$''. Under Wilcoxon rank sum test with significance level of $0.05$, an algorithm is marked with underline (in bold) if it is significantly worse (better) than RCO-SAHiD.}
\label{table:sahid-beijing-ave}
\begin{center}
\begin{tabular}{@{}ccccccccccccccccc@{}}
\toprule
Name & $|V|$ & $|E|$ & $|T|$ & $\zeta$  & \multicolumn{ 2}{c}{RDG-MAENS} & \multicolumn{ 2}{c}{VNS} & \multicolumn{ 2}{c}{TSA1} & \multicolumn{ 2}{c}{UHGS}& \multicolumn{ 2}{c}{SAHiD} &\multicolumn{ 2}{c}{RCO-SAHiD} \\
\cmidrule(lr){6-7} \cmidrule(lr){8-9} \cmidrule(lr){10-11} \cmidrule(lr){12-13}\cmidrule(lr){14-15}\cmidrule(lr){16-17}

 & &  &  &  &    Mean &        Std &    Mean &        Std &    Mean &Std &    Mean &        Std &    Mean &        Std &    Mean &        Std \\
\midrule
 {Beijing}-1 &       2820 &       3584 &        358 &          7 &{\underline{829406}} &      12688 & { {\underline{782415}}} &       4452 & {\underline{829132}} &       6340 & {\bf{760578}$\dagger$} &       0 & {\underline{784727}} &       5591 & 770199   &     3178 \\

 {Beijing}-2 &       2820 &       3584 &        717 &         11 &{\underline{1337954}} &      18939 & {\underline{1192292}} &      10196 & {\underline{1401363}} &      25378 & {\bf{1132987}$\dagger$} &       1638 &{\underline{1183955}} &       8431 & 1163978   &     6258 \\

 {Beijing}-3 &       2820 &       3584 &       1075 &         18 & {\underline{1847922}} &      33258 & {\underline{1618484}} &      11888 & {\underline{1709279}} &      14801 &{\bf{1542405}$\dagger$} &       3801 &{\underline{1605846}} &       9231 & 1577027   &     6798 \\

 {Beijing}-4 &       2820 &       3584 &       1434 &         23 & {\underline{2193399}} &      34159 & {\underline{1953892}} &      16746 & {\underline{2070885}} &      14532 & {\bf{1847355}$\dagger$} &      5571 &{\underline{1936994}} &      11694 & 1896581   &     8411 \\

 {Beijing}-5 &       2820 &       3584 &       1792 &         30 & {\underline{2639458}} &      32481 & {\underline{2335915}} &      23040 & {\underline{2440319}} &      26726 & {\bf{2210443}$\dagger$} &      5638 &{\underline{2298630}} &      16879 & 2255386   &     8316 \\

 {Beijing}-6 &       2820 &       3584 &       2151 &         36 & {\underline{3047295}} &      41112 & {\underline{2743677}} &      18024 & {\underline{2814735}} &      22018 & {\bf{2571748}$\dagger$} &       6003 &{\underline{2707500}} &      18433 & 2650420   &     9621 \\

 {Beijing}-7 &       2820 &       3584 &       2509 &         41 & {\underline{3388263}} &      26081 & {\underline{3063813}} &      25226 & {\underline{3186240}} &      22426 & {\bf{2871881}$\dagger$} &       10590 &{\underline{3038157}} &      15658 &2952809   &    14474 \\

 {Beijing}-8 &       2820 &       3584 &       2868 &         47 & {\underline{3697025}} &      44951 & {\underline{3366215}} &      24686 & {\underline{3456037}} &      22381 & {\bf{3150688}$\dagger$} &       7879 &{\underline{3313590}} &      21925 & 3233296   &    15953 \\

 {Beijing}-9 &       2820 &       3584 &       3226 &         52 & {\underline{4061793}} &      49504 & {\underline{3723830}} &      45148 & {\underline{3943883}} &      37089 & {\bf{3485819}$\dagger$} &      10731 &{\underline{3684250}} &      32404 &3575671   &    15372 \\

{Beijing}-10 &       2820 &       3584 &       3584 &         58 & {\underline{4353966}} &      51063 & {\underline{4040694}} &      27384 & {\underline{4103532}} &      15501 & {\bf{3785520}$\dagger$} &       11830&{\underline{4004310}} &      29488 &3884308   &    16206 \\[5pt]

    W-D-L &            &            &            &            &     10-0-0 &            &     10-0-0 &            &     10-0-0 &    & 0-0-10&         &     10-0-0 &            &            &            \\
\bottomrule
\end{tabular}
\end{center}
\end{table*}



\begin{table*}[!ht]
\footnotesize
  \caption{The best total cost of the compared algorithms on the Beijing dataset. For each instance, the minimal total cost is marked with ``$\dagger$''. An algorithm is marked with underline (in bold) if its total cost is greater (smaller) than RCO-SAHiD.}
  \label{table:sahid-beijing-best}
\begin{center}
\begin{tabular}{ccccccc}
\toprule
      Name &      RDG-MAENS &        VNS &       TSA1 & UHGS & SAHiD & RCO-SAHiD \\
\midrule
 {Beijing}-1 & \underline{812647}    & \underline{774502}   {} & \underline{ 813907}    &  \bf{760578}$\dagger$   & \underline{775523} & {765538} \\

 {Beijing}-2 & \underline{1303570}    & \underline{1168190}    & \underline{ 1353567}   &  \bf{1129810}$\dagger$   & \underline{1167480} & {1148259} \\

 {Beijing}-3 & \underline{1777852}    & \underline{1591540}    & \underline{1678224}    &  \bf{1534878}$\dagger$   & \underline{1586180} & {1563874}  \\

 {Beijing}-4 & \underline{2126151}    & \underline{1920330}    & \underline{2053938}    &  \bf{1836866}$\dagger$   & \underline{1910880} & {1879617 } \\

 {Beijing}-5 & \underline {2581910}    & \underline{2293120}    & \underline{ 2396483}   &  \bf{2199275}$\dagger$   & \underline{2273080} &  2234352  \\

 {Beijing}-6 & \underline{2968102}    & \underline{2705060}    & \underline{2774161}    &  \bf{2561113}$\dagger$   & \underline{2664510} &   2632250  \\

 {Beijing}-7 & \underline{3331900}    & \underline{3015790}    & \underline{3147294}    &  \bf{2851602}$\dagger$   & \underline{3013590} &   2925015  \\

 {Beijing}-8 & \underline{3584696}    & \underline{3323850}    & \underline{3415275}    &  \bf{3136727}$\dagger$   & \underline{3283530} &  3203032  \\

 {Beijing}-9 & \underline{3934270}    & \underline{3653630}    & \underline{3890129}    &  \bf{3462953}$\dagger$   & \underline{3621490} &  3541842  \\

{Beijing}-10 & \underline{4206005}    & \underline{ 4002040}   & \underline{4066188}    &  \bf{3765614}$\dagger$   & \underline{3935540} &   3852428  \\[5pt]

      Mean &   2662710.3 &    2444805.2 &    2558916.6 & 2323941.6  &2423180.3 &  2374620.7 \\

        G-E-S &          10-0-0 &          10-0-0 &          10-0-0 &          0-0-10 &        10-0-0 &          \\
\bottomrule
\end{tabular}
\end{center}
\end{table*}

To further demonstrate the efficacy of embedding RCO, Fig. \ref{fig:curves-hb} shows the convergence curves of RCO-SAHiD and SAHiD over the 25 independent runs on the Hefei-1, Hefei-10, Beijing-1 and Beijing-10 instances, where the $x$-axis is the computational time in seconds, and the $y$-axis is the average total cost of the best-so-far solution. These four instances are selected as the representative instances of the corresponding datasets, and similar patterns are observed on other instances.

From the figure, one can see that the convergence curves of RCO-SAHiD are almost always below that of SAHiD, and there is a decent gap between the two convergence curves. The only exception is the Hefei-10 instance, for which SAHiD converged slightly better than RCO-SAHiD. This is also consistent with Table \ref{table:sahid-hefei-ave}, which shows that Hefei-10 is the only instance where there is no statistical difference between RCO-SAHiD and SAHiD.

\begin{figure*}[!ht]
	\centering
    \includegraphics[width=0.45\textwidth]{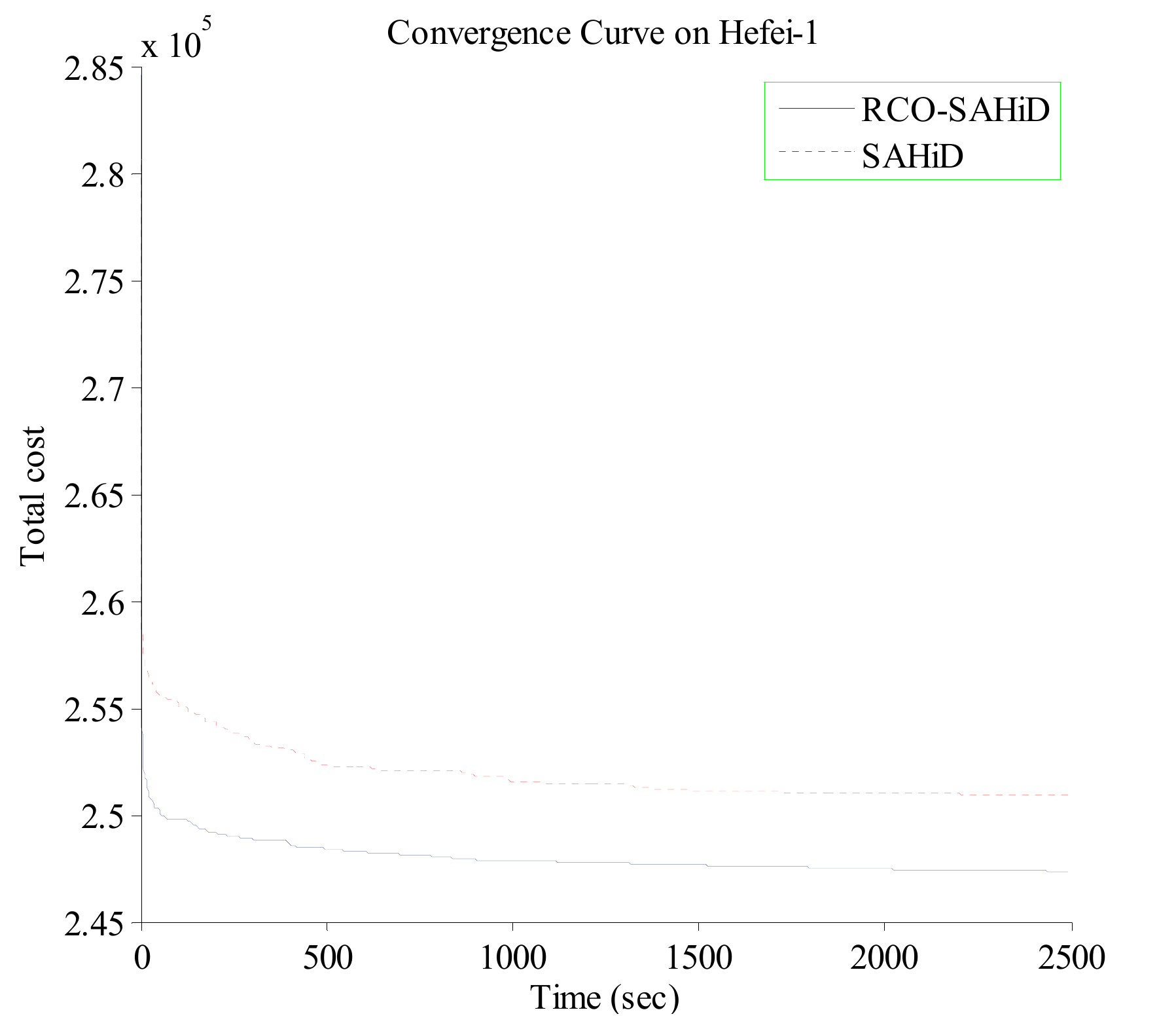}
    \includegraphics[width=0.45\textwidth]{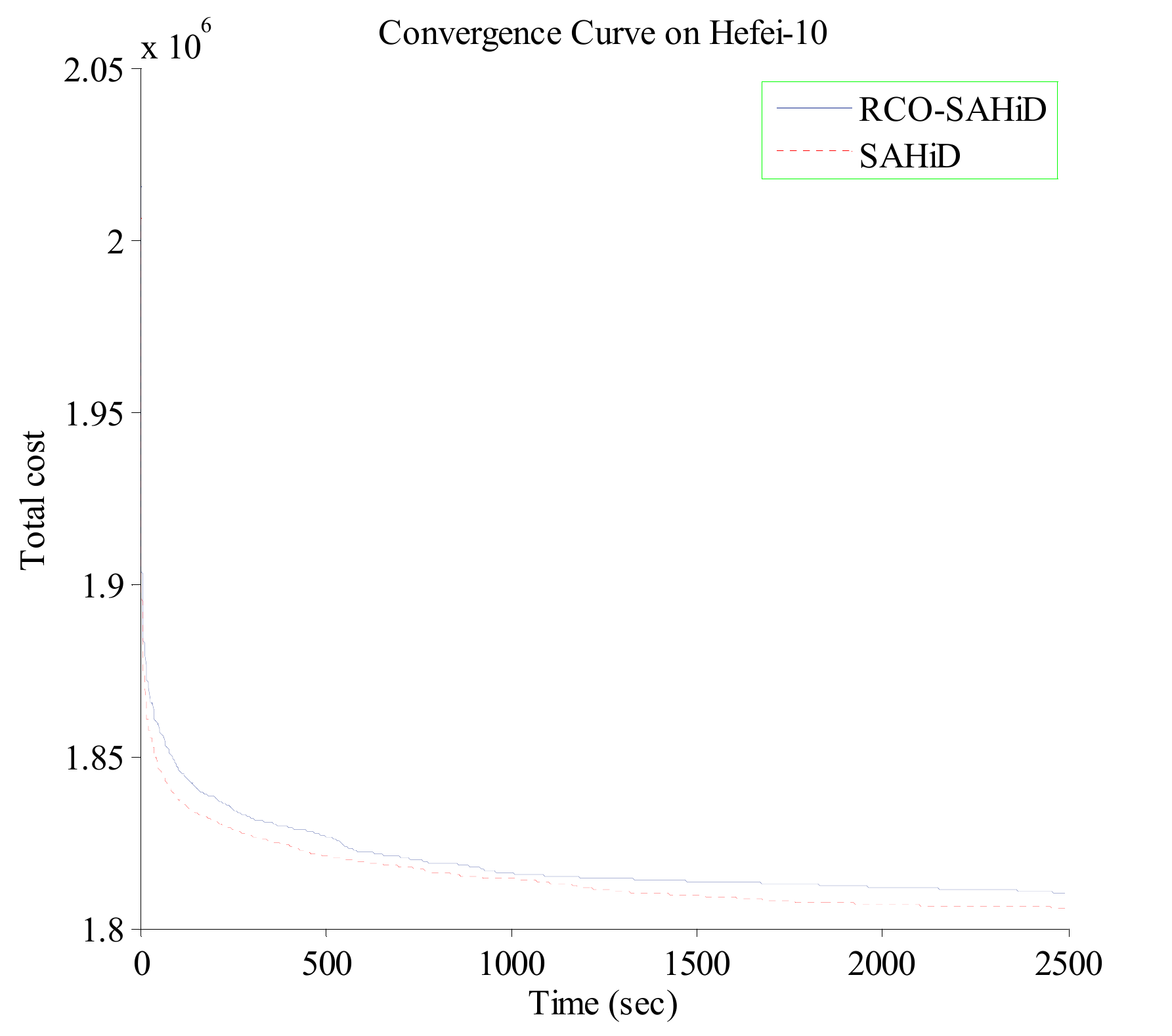}
    \includegraphics[width=0.45\textwidth]{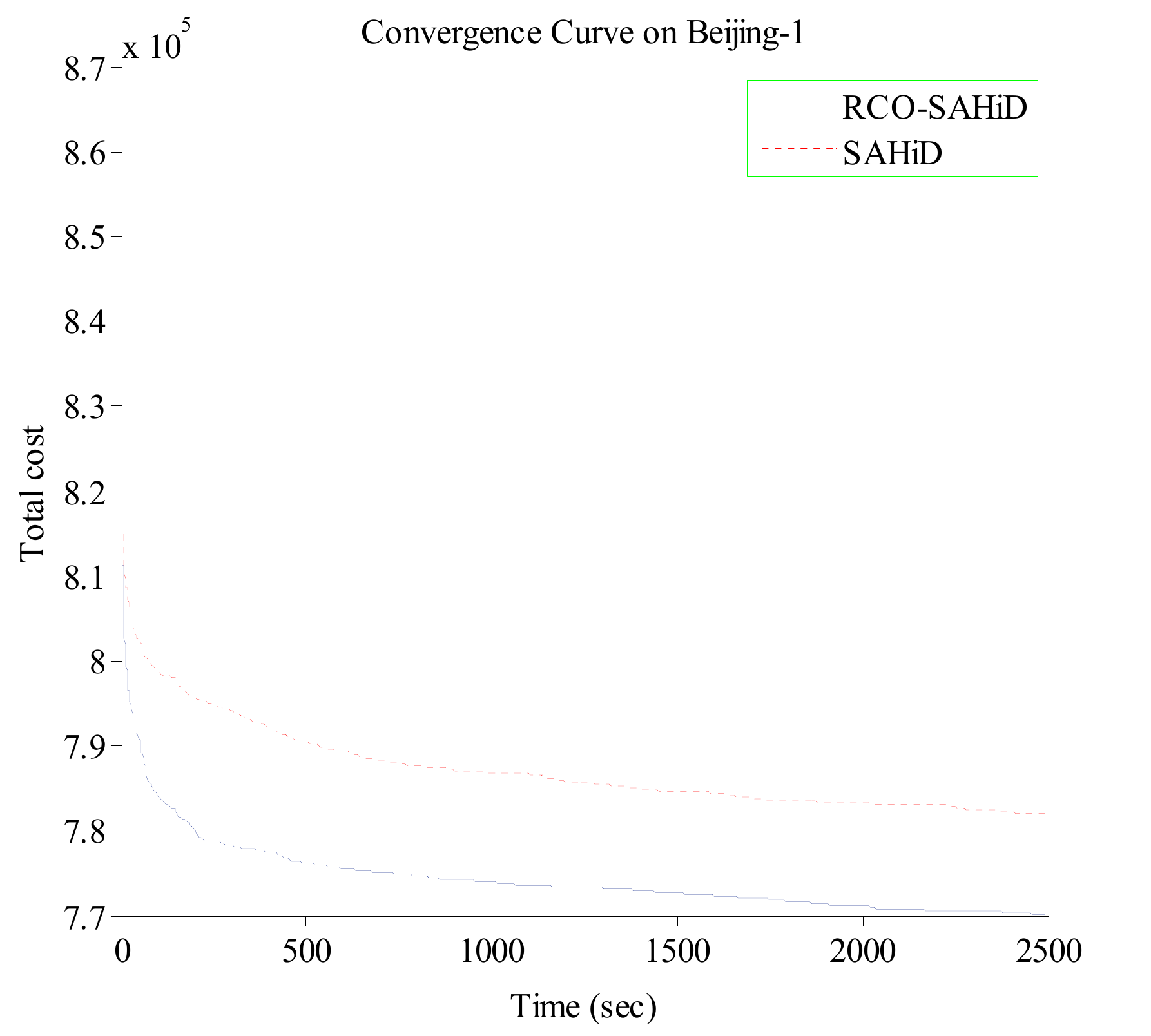}
    \includegraphics[width=0.45\textwidth]{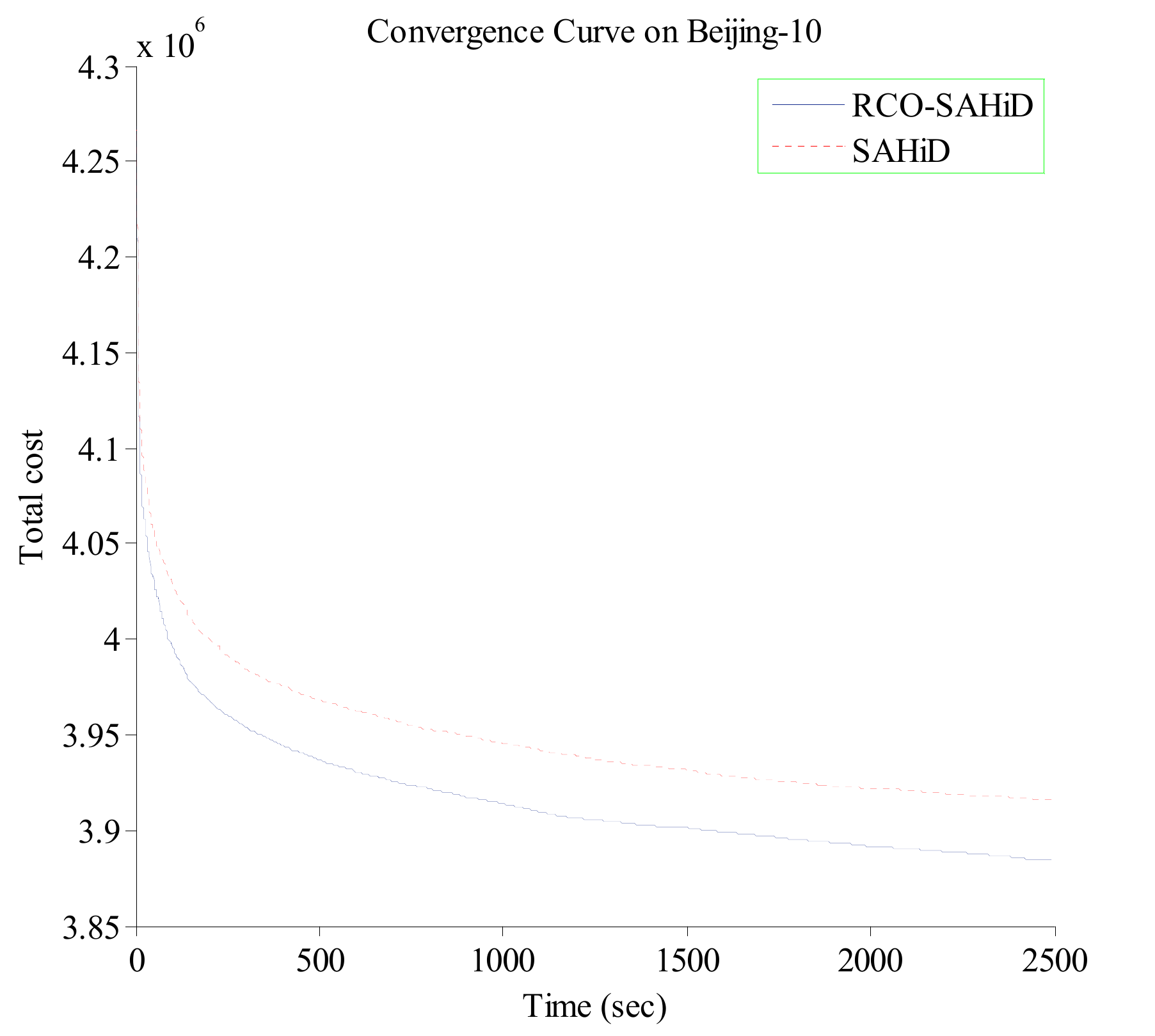}
    \caption{Convergence curves of RCO-SAHiD and SAHiD on the Hefei-1, Hefei-10, Beijing-1 and Beijing-10 instances.}
    \label{fig:curves-hb}
\end{figure*}

In summary, the results in Experiment 2 clearly demonstrate the effectiveness of the RCO operator in improving the performance of SAHiD. Although RCO-SAHiD did not perform so well as UHGS, this is mainly due to the superiority of UHGS over SAHiD in terms of search capability, rather than the problem decomposition. Since UHGS is not a divide-and-conquer approach, we expect that embedding RCO into UHGS can further improve its performance.


\subsection{Results on Experiment 3}

Experiment 3 is to compare RCO-SAHiD with SAHiD \cite{tang2017scalable}, UHGS \cite{vidal2017node}, Fast-CARP \cite{wohlk2018fast} and PS \cite{zbib2017vriants} on 12 largest KW instances. RCO-SAHiD, SAHiD and UHGS were ran 25 times independently. The runtime of $81$ seconds per $1000$ nodes is set for the compared algorithms on each instance. The results of Fast-CARP and PS are copied from the original literatures (\cite{wohlk2018fast} and \cite{zbib2017vriants}), as their codes are not available.

Compared with Experiment 2, Experiment 3 has a much larger problem size and much tighter time budget. For example, in K1\_g-2, there are 8556 tasks, while only $81 \times 11640 / 1000 = 943$ seconds is allowed. In other words, Experiment 3 has a strong requirement for an algorithm to search effectively in a huge search space within a very limited time budget.

Table \ref{table:sahid-KW-ave} shows the average performance of the compared algorithms on the 12 KW instances. For each instance, the minimal mean total cost is marked with ``$\dagger$''. Under Wilcoxon rank sum test with significance level of $0.05$, if an algorithm is significantly worse (better) than RCO-SAHiD, then its result is marked with underline (in bold).
\begin{table*}[!ht]
\footnotesize
\caption{The average performance over 25 independent runs of the compared algorithms on the KW dataset. For each instance, the minimal mean total cost is marked with ``$\dagger$''. Under Wilcoxon rank sum test with significance level of $0.05$, an algorithm is marked with underline (in bold) if it is significantly worse (better) than RCO-SAHiD.}
\label{table:sahid-KW-ave}
\begin{center}
\begin{tabular}{@{}ccccccccccccc@{}}
\toprule
Name & $|V|$ & $|E|$ & $|T|$ & $Q$  &  \multicolumn{ 2}{c}{UHGS}& \multicolumn{ 2}{c}{SAHiD} &\multicolumn{ 2}{c}{RCO-SAHiD} \\
\cmidrule(lr){6-7}  \cmidrule(lr){8-9} \cmidrule(lr){10-11}

 & &  &  &  &   Mean &Std &    Mean &        Std &    Mean &        Std &   \\
\midrule
    K1\_g-2 & 11640 & 12675 & 8566  & \multicolumn{1}{c}{48000} & \textbf{6361639}$\dagger$ & 62021 & \underline{6555725} & 20127 & 6464425 & 28374 \\
    K1\_g-6  & 11640 & 12675 & 8566  & \multicolumn{1}{c}{168000} & \textbf{3593703}$\dagger$ & 23857 & \underline{3806616} & 16567 & 3785654 & 15569 \\
    K2\_g-2 & 11636 & 12671 & 8563  & \multicolumn{1}{c}{48000} & \textbf{6173748}$\dagger$ & 69412 & \underline{636148}6 & 25149 & 6268937 & 27169 \\
    K2\_g-4  & 11636 & 12671 & 8563  & \multicolumn{1}{c}{96000} & \textbf{4353692}$\dagger$ & 44890 & \underline{4484168} & 26174 & 4463470 & 22877 \\
    K5\_g-2 & 11405 & 12435 & 8267  & \multicolumn{1}{c}{48000} & \textbf{5931093}$\dagger$ & 70015 & \underline{6143480} & 24601 & 6066358 & 25310 \\
    K5\_g-6 & 11405 & 12435 & 8267  & \multicolumn{1}{c}{168000} & \textbf{3447989}$\dagger$ & 26460 & \underline{3652255} & 18995 & 3637475 & 21134 \\
    O1\_g-4 & 10283 & 11863 & 8581  & \multicolumn{1}{c}{96000} & \textbf{3291833}$\dagger$ & 28719 & \underline{3428522} & 12159 & 3380808 & 11764 \\
    O1\_g-6 & 10283 & 11863 & 8581  & \multicolumn{1}{c}{168000} & \textbf{2700608}$\dagger$ & 24040 & \underline{2819667} & 13203 & 2800130 & 15190 \\
    O1\_p-2 & 9957  & 11492 & 8220  & \multicolumn{1}{c}{48000} & \textbf{2515957}$\dagger$ & 23390 & \underline{2622827} & 11350 & 2611434 & 11455 \\
    O1\_p-4 & 9957  & 11492 & 8220  & \multicolumn{1}{c}{96000} & \textbf{2210256}$\dagger$ & 18760 & \underline{2314768} & 12627 & 2300794 & 14922 \\
    O6\_g-2 & 9563  & 11073 & 7831  & \multicolumn{1}{c}{48000} & \textbf{3511678}$\dagger$ & 35021 & \underline{3657354} & 12754 & 3595682 & 13912 \\
    O6\_g-6 & 9563  & 11073 & 7831  & \multicolumn{1}{c}{168000} & \textbf{2269116}$\dagger$ & 15096 & \underline{2371623} & 13594 & 2361511 & 11524    \\[5pt]

    W-D-L &            &            &            &            &     0-0-12 &         &    12-0-0 &               \\
\bottomrule
\end{tabular}
\end{center}
\end{table*}

From Table \ref{table:sahid-KW-ave}, one can see that RCO-SAHiD statistically significantly outperformed SAHiD on all the 12 KW instances. It is consistent with the results on the Hefei and Beijing datasets, which demonstrates the effectiveness of the proposed RCO operator.
Again, UHGS performed the best on the KW datasets. However, we observed that it is much slower than our algorithm. Specifically, even the initialisation stage can take much longer time than the given time budget (e.g. $3774.5$ seconds for the O1\_p-4 instance, while the given time budget is $804.2$ seconds).

Table \ref{table:sahid-KW-best} shows the best performance of the compared algorithms on the KW dataset, where the minimal total cost of each instance is marked with ``$\dagger$'', and is marked with underline (in bold) if it is greater (smaller) than RCO-SAHiD. For Fast-CARP and PS, the results are obtained from \cite{wohlk2018fast} directly. Since their results were obtained by a single run after sophisticated parameter tuning  (i.e. Fast-CARP), or the best ones obtained by 105 runs with different evaluation criteria and degrees of randomization (i.e. PS), we treated them as the best performance in the comparison.
The patterns shown in Table \ref{table:sahid-KW-best} are consistent with those in Table \ref{table:sahid-KW-ave}.
In addition, Fast-CARP performed slightly better than RCO-SAHiD on some KW instances. However, it was very carefully tuned, by testing a large number of combinations of parameters (about 200 combinations). On the other hand, SAHiD is almost parameter-free, and is easier to use in practice.


\begin{table*}[!ht]
\footnotesize
  \caption{The best total cost of the compared algorithms on the Beijing dataset. For each instance, the minimal total cost is marked with ``$\dagger$''. An algorithm is marked with underline (in bold) if its total cost is greater (smaller) than RCO-SAHiD.}
  \label{table:sahid-KW-best}
\begin{center}
\begin{tabular}{cccccc}
\toprule
      Name &      PS &       Fast-CARP &     UHGS & SAHiD & RCO-SAHiD \\
\midrule
  K1\_g-2 & \underline{7549094} & \underline{6501210} & \textbf{6274277}$\dagger$ & \underline{6509006} & 6405640 \\
    K1\_g-6  & \underline{4886345} & \textbf{3739724} & \textbf{3541665}$\dagger$ & \underline{3771667} & 3760040 \\
    K2\_g-2 & \underline{7434491} & \underline{6249733} & \textbf{6033790$\dagger$} & \underline{6319036} & 6224123 \\
    K2\_g-4  & \underline{5629112} & \underline{4434203} & \textbf{4293975}$\dagger$ & \underline{4445174} & 4416990 \\
    K5\_g-2 & \underline{7136324} & \underline{6031579} & \textbf{5822825$\dagger$} & \underline{6081146} & 6008824 \\
    K5\_g-6 & \underline{4707990} & \textbf{3578627} & \textbf{3406571}$\dagger$ & \underline{3609300} & 3596594 \\
    O1\_g-4 & \underline{4173408} & \textbf{3278666} & \textbf{3260611}$\dagger$ & \underline{3404825} & 3364902 \\
    O1\_g-6 & \underline{3596504} & \textbf{2724848} & \textbf{2664709}$\dagger$ & \underline{2794956} & 2766860 \\
    O1\_p-2 & \underline{3465916} & \textbf{2509047} & \textbf{2484393}$\dagger$ & \underline{2600948} & 2585479 \\
    O1\_p-4 & \underline{3089185} & \textbf{2194629} & \textbf{2173807}$\dagger$ & \underline{2285241} & 2268133 \\
    O6\_g-2 & \underline{4345961} & \textbf{3531246} & \textbf{3459355}$\dagger$ & \underline{3639059} & 3574446 \\
    O6\_g-6 & \underline{3100213} & \textbf{2276829} & \textbf{2239678}$\dagger$ & \underline{2352559} & 2343073  \\\\
    Mean  & 4926211.9  & 3920861.8  & 3804638.0  & 3984409.8  & 3942925.3  \\
    G-E-S &          12-0-0 &          4-0-8 &          0-0-12 &          12-0-0 &           \\
\bottomrule
\end{tabular}
\end{center}
\end{table*}
Fig. \ref{fig:curves-kw} shows the convergence curves of RCO-SAHiD and SAHiD over 25 independent runs on four representative KW instances (K1\_g-2, K1\_g-6, O6\_g-2 and O6\_g-6), where the $x$-axis is the computational time in seconds, and the $y$-axis is the average total cost of the best-so-far solution. These four instances are selected as the representative instances of the corresponding datasets, and similar patterns are observed on other instances.

From the figure, one can see that the convergence curves of RCO-SAHiD are always significantly below that of SAHiD, and there is an apparent gap between the two convergence curves. This indicates that no matter when to stop the search, RCO-SAHiD will always provide a better solution than SAHiD.

\begin{figure*}[!ht]
	\centering
    \includegraphics[width=0.45\textwidth]{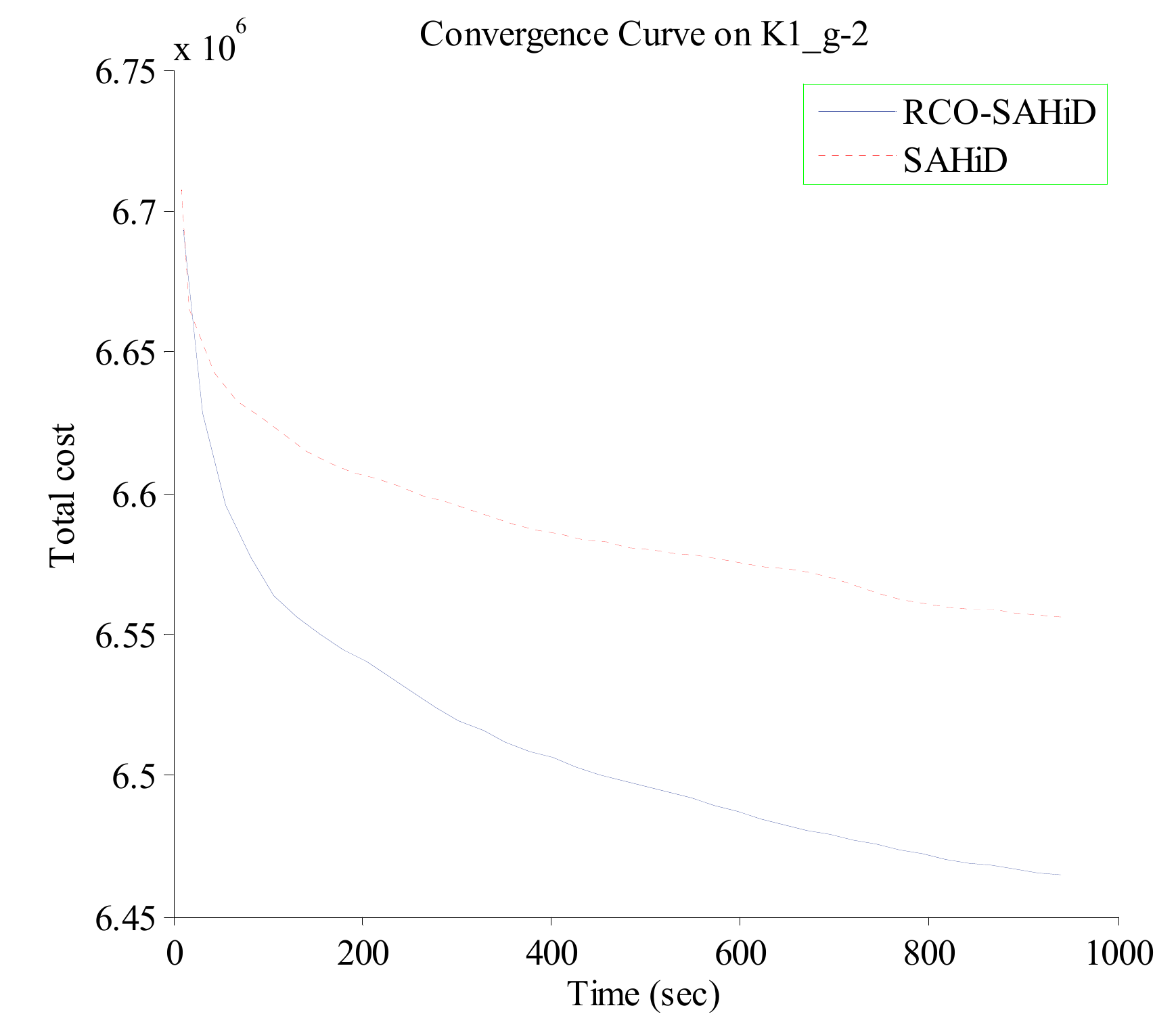}
    \includegraphics[width=0.45\textwidth]{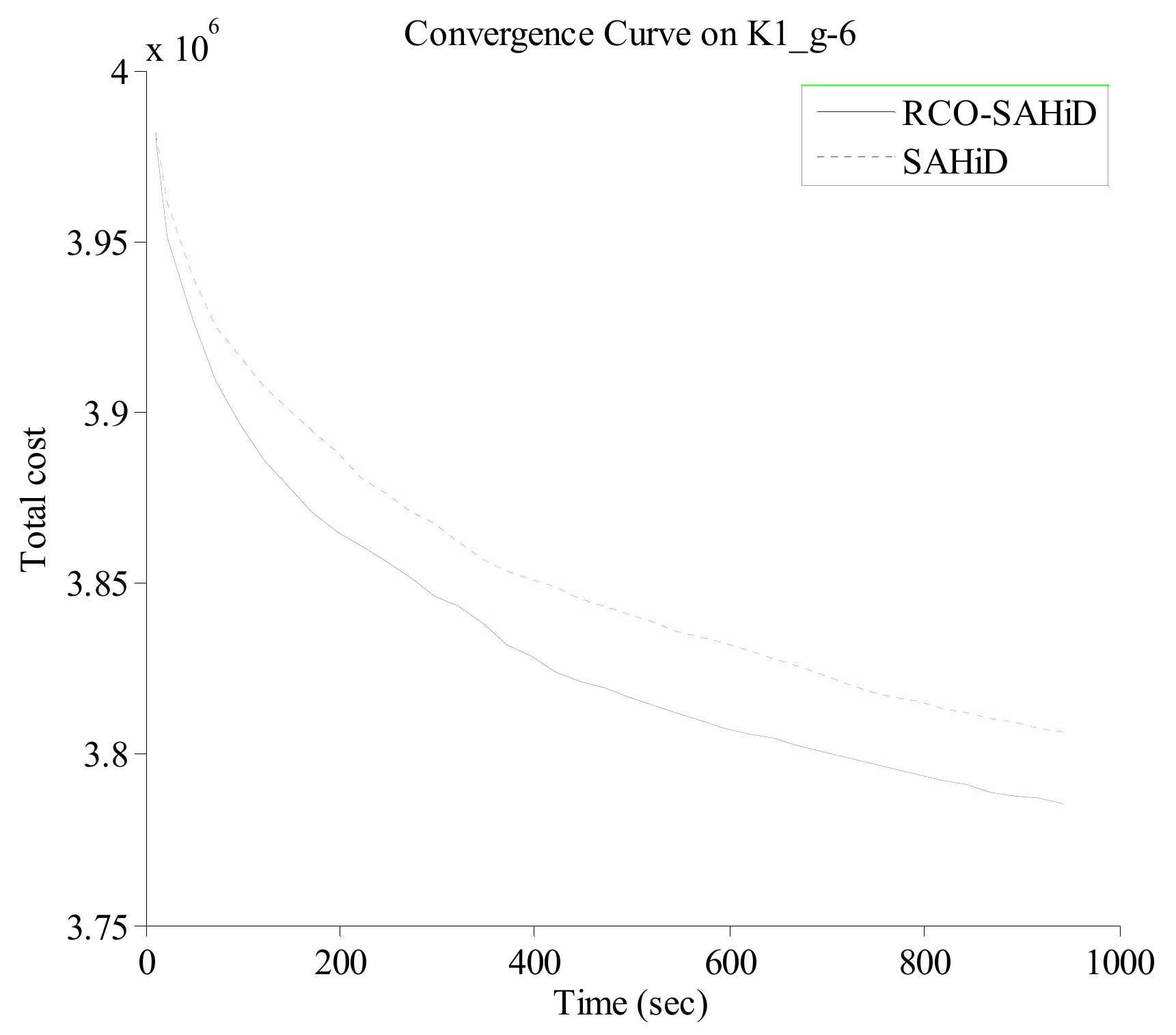}
    \includegraphics[width=0.45\textwidth]{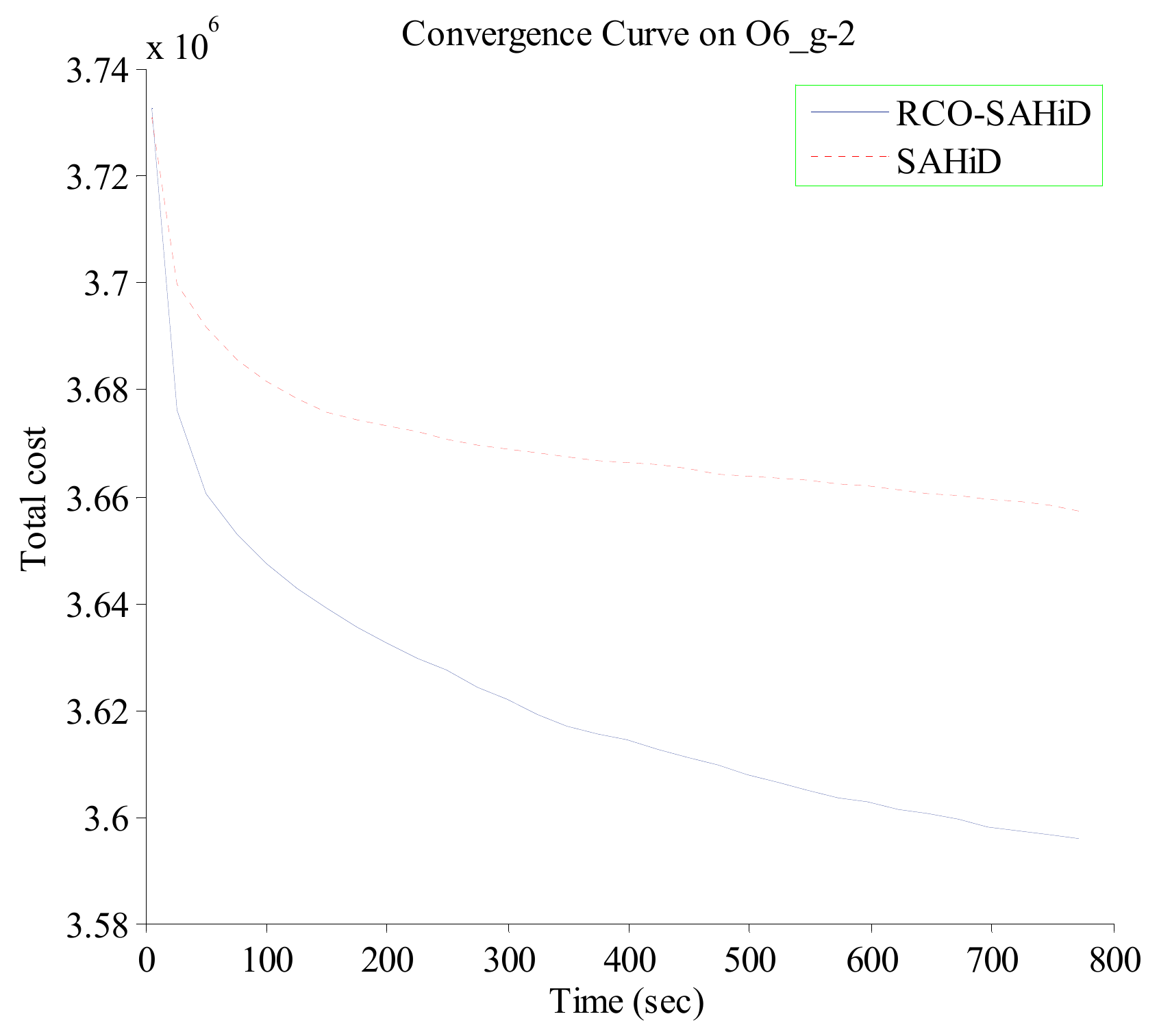}
    \includegraphics[width=0.45\textwidth]{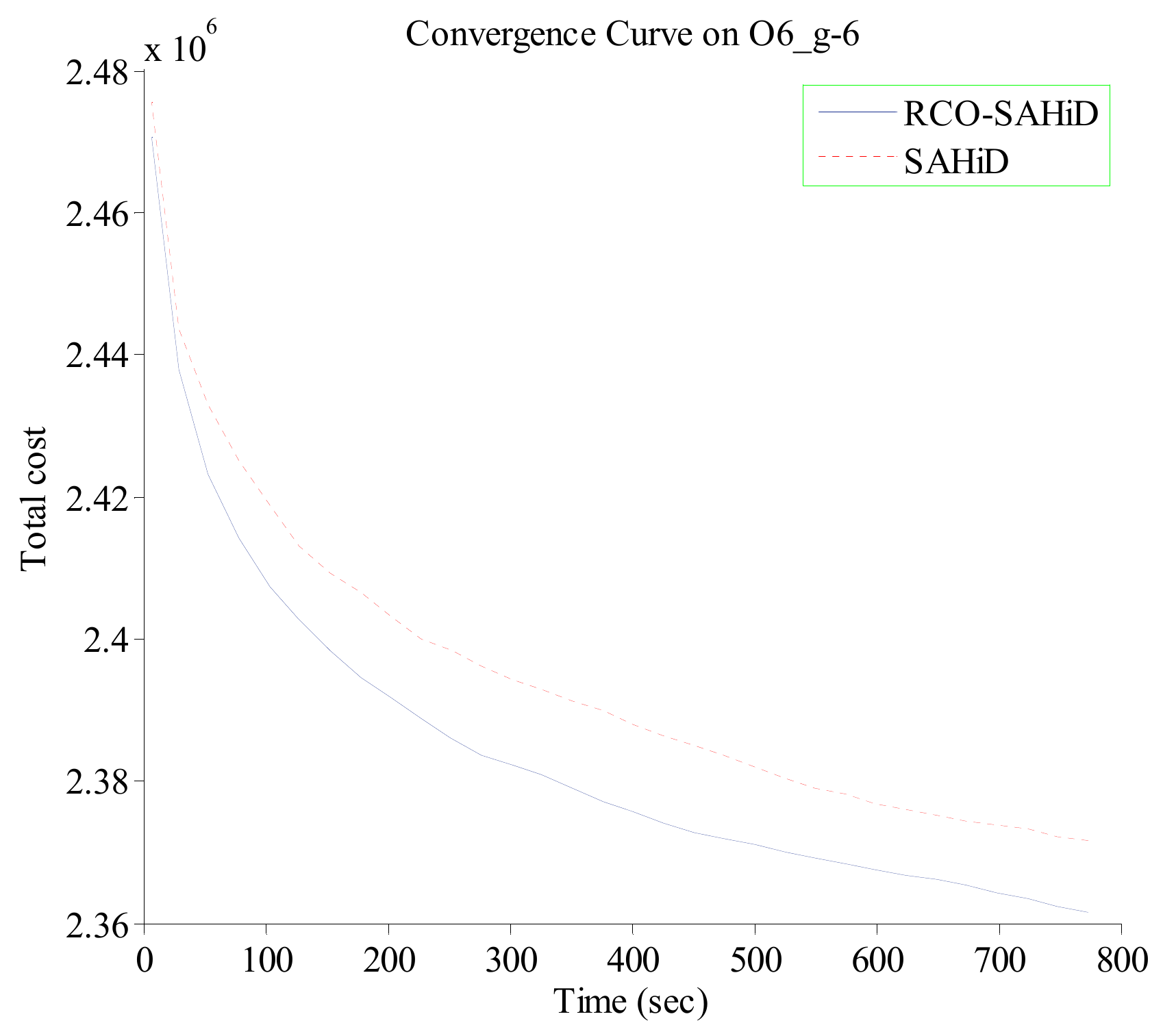}
    \caption{Convergence curves of RCO-SAHiD and SAHiD on the K1\_g-2, K1\_g-6, O6\_g-2 and O6\_g-6 instances.}
    \label{fig:curves-kw}
\end{figure*}

In summary, the results in Experiment 3 clearly demonstrates the effectiveness of the proposed RCO operator in improving the SAHiD algorithm on large instances. Although RCO-SAHiD did not manage to outperform the state-of-the-art UHGS and Fast-CARP, due to the limited performance of the embedded SAHiD, it was still faster and less space demanding than UHGS, and less parameter sentitive than Fast-CARP.

\subsection{Scalability}

To investigate how the effectiveness of the RCO operator is affected by the problem size, we plot the relative performance of RCO-SAHiD to SAHiD on different problem sizes.
Fig. \ref{fig:scalability} shows the relative performance of RCO-SAHiD to SAHiD (ratio between their performances) versus the number of tasks on all the tested instances, where the $x$-axis is the number of tasks, and the $y$-axis is relative performance. 

From Fig. \ref{fig:scalability}, one can see that embedding RCO can improve the performance of SAHiD (the ratio is smaller than 1) on almost all the instances except one Hefei instance. For the Beijing instances, the advantage of RCO becomes more obvious as the problem size increases. For the Hefei instances, on the other hand, RCO becomes less effective with the increase of the problem size. For the EGLG and KW instances, there is no particular trend, since they have very similar problem sizes.
Overall, RCO can almost always lead to improvement.

To give a further analysis on the poor performance of RCO on the Hefei dataset, we investigated the topology of the original network on Hefei-10 by computing the link rank of each pair of tasks. We observed that Hefei-10 contains less high-rank links than other instances. Such link distribution decreases the effectiveness of RCO in catching poor links. As a result, RCO-SAHiD can hardly obtain high-quality solution to Hefei-10, and showed poor performance.


\begin{figure*}[!ht]
	\centering
    \includegraphics[width=0.45\textwidth]{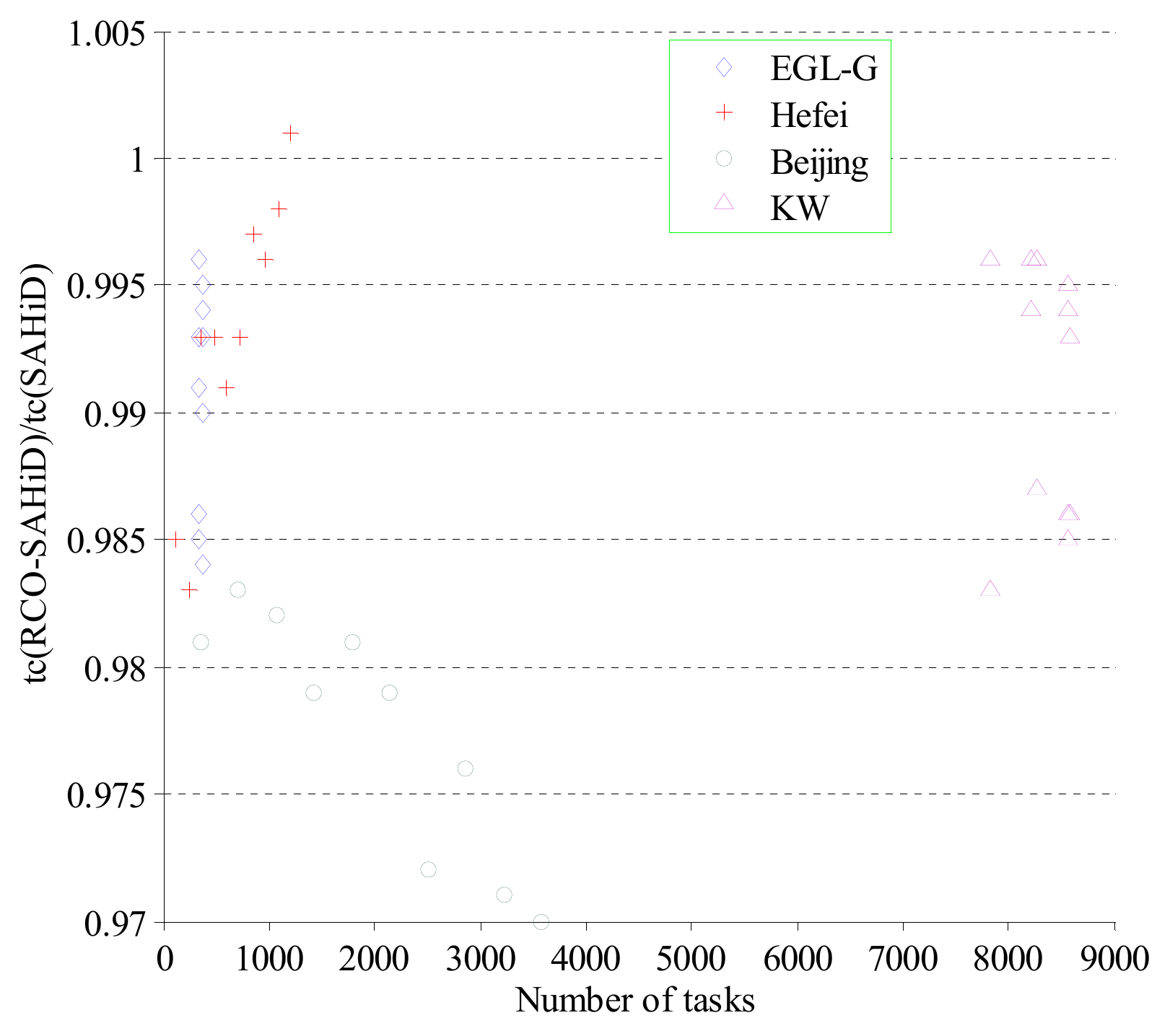}
    \caption{The relative performance of RCO-SAHiD to SAHiD versus the number of tasks on the tested datasets.}
    \label{fig:scalability}
\end{figure*}

\section{Conclusions and Future Work}
\label{sec:conc}

LSCARP is a hot topic of research on CARP, and a number of competitive algorithms have been proposed for solving it. 
In this paper, to better decompose the problem, a novel operator named the Route Cutting Off (RCO) operator is proposed for splitting the routes of the best-so-far solutions during the search process. The RCO operator is based on the idea that within the same route of the best-so-far solution, some links between tasks are good, and some others may be poor. We designed a task rank matrix, based on which the RCO operator classifies the links into good links and poor links. Then, it cuts off the two types of links with certain probabilities, in order to provide a better task subset for clustering, and thus lead to a better decomposition.

To verify the effectiveness of the proposed RCO operator, we propose two divide-and-conquer algorithms, namely RCO-RDG-MAENS and RCO-SAHiD, which are obtained by embedding the RCO operator into RDG-MAENS \cite{mei2014cooperative} and SAHiD \cite{tang2017scalable}, respectively.
The experimental results clearly showed that the RCO operator managed to improve the performance of both RDG-MAENS and SAHiD, especially when the problem size is very large and the time budget is very limited. Note that the RCO operator is very flexible, and can be easily embedded into any divide-and-conquer approach that decomposes the problem by clustering the tasks.

The possible future directions can be as follows. First, the current link classification can be improved. Currently, we simply use the average task rank of the solution as a threshold, and consider a link to be good if its rank is lower than the average task rank. The threshold is set in a rather arbitrary way, and may be improved in the future. Second, currently each route can have at most two links (one good link and one poor link) cut off during RCO. However, it may be more desirable to cut off more poor links than the good links, especially for long routes. In the future, we will consider adaptive schemes for deciding the number of cut-off links. Third, we will improve the robustness of the RCO operator in different graph topology.

\section*{Acknowledgment}

This work was supported by Anhui Provincial Natural Science Foundation (Nos. 1808085MF173, 1908085MF195), Natural Science Key Research Project for Higher Education Institutions of Anhui Province(Nos. KJ2016A438, KJ2017A352), the National Key R \& D Program of China under Grant 2017YFC1601800, and the National Natural Science Foundation of China (No. 61673194).

\section*{References}

\bibliographystyle{plain}
\bibliography{Bibliography}

\begin{thebibliography}{10}

\bibitem{baldacci2010exact}
R.~Baldacci and V.~Maniezzo.
\newblock {Exact methods based on node-routing formulations for undirected
  arc-routing problems}.
\newblock {\em Networks}, 47(1):52--60, 2010.

\bibitem{bartolini2013improved}
E.~Bartolini, J.-F. Cordeau, and G.~Laporte.
\newblock {Improved lower bounds and exact algorithm for the capacitated arc
  routing problem}.
\newblock {\em Mathematical Programming Series B}, 137(1--2):409--452, 2013.

\bibitem{belenguer2003cutting}
J.M. Belenguer and E.~Benavent.
\newblock {A cutting plane algorithm for the capacitated arc routing problem}.
\newblock {\em Computers and Operations Research}, 30(5):705--728, 2003.

\bibitem{benavent1992capacitated}
E.~Benavent, V.~Campos, A.~Corber{\'a}n, and E.~Mota.
\newblock {The capacitated arc routing problem: lower bounds}.
\newblock {\em Networks}, 22(7):669--690, 1992.

\bibitem{beullens2003guided}
P.~Beullens, L.~Muyldermans, D.~Cattrysse, and D.~Van~Oudheusden.
\newblock {A guided local search heuristic for the capacitated arc routing
  problem}.
\newblock {\em European Journal of Operational Research}, 147(3):629--643,
  2003.

\bibitem{schlebusch2012cut}
C.~Bode and S.~Irnich.
\newblock {Cut-first branch-and-price-second for the capacitated arc-routing
  problem}.
\newblock {\em Operations research}, 60(5):1167--1182 ., 2012.

\bibitem{brandao2008deterministic}
J.~Brand{\~a}o and R.~Eglese.
\newblock {A deterministic tabu search algorithm for the capacitated arc
  routing problem}.
\newblock {\em Computers and Operations Research}, 35(4):1112--1126, 2008.

\bibitem{chen2018phased}
Y.~Chen and J.K. Hao.
\newblock {Two phased hybrid local search for the periodic capacitated arc
  routing problem}.
\newblock {\em European Journal of Operational Research}, pages 1--24, 2018.

\bibitem{dearmon1981comparison}
J.S. DeArmon.
\newblock {A comparison of heuristics for the capacitated Chinese postman
  problem}.
\newblock Master's thesis, University of Maryland, 1981.

\bibitem{dijkstra1959note}
E.W. Dijkstra.
\newblock {A note on two problems in connection with graphs}.
\newblock {\em Numerische mathematik}, 1(1):269--271, 1959.

\bibitem{eglese1996tabu}
R.W. Eglese and L.Y.O. Li.
\newblock {A tabu search based heuristic for arc routing with a capacity
  constraint and time deadline}.
\newblock {\em Meta-Heuristics: Theory $\&$ Applications, Kluwer Academic
  Publishers, Boston}, pages 633--650, 1996.

\bibitem{eiselt1995arc}
H.A. Eiselt and M.~Gendreau.
\newblock {Arc routing problems, part II: the rural postman problem}.
\newblock {\em Operations Research}, 43(3):399--414, 1995.

\bibitem{feng2010towards}
L.~Feng, Y.S. Ong, Q.H. Nguyen, and A.H. Tan.
\newblock {Towards probabilistic memetic algorithm: An initial study on
  capacitated arc routing problem}.
\newblock In {\em Proceedings of the 2010 IEEE Congress on Evolutionary
  Computation}, pages 18--23, 2010.

\bibitem{goh2009competitive}
C.K. Goh and K.C. Tan.
\newblock {A competitive-cooperative coevolutionary paradigm for dynamic
  multiobjective optimization}.
\newblock {\em IEEE Transactions on Evolutionary Computation}, 13(1):103--127,
  2009.

\bibitem{golden1981capacitated}
B.L. Golden and R.T. Wong.
\newblock {Capacitated arc routing problems}.
\newblock {\em Networks}, 11(3):305--316, 1981.

\bibitem{handa2006robustmagz}
H.~Handa, L.~Chapman, and X.~Yao.
\newblock {Robust route optimization for gritting/salting trucks: a CERCIA
  experience}.
\newblock {\em IEEE Computational Intelligence Magazine}, 1(1):6--9, 2006.

\bibitem{kiilerich2017new}
L.~Kiilerich and W{\o}hlk S.
\newblock {New large-scale data instances for CARP and new variations of CARP}.
\newblock {\em INFOR: Information Systems and Operational Research},
  56(1):1--32, 2018.

\bibitem{lacomme2004competitive}
P.~Lacomme, C.~Prins, and W.~Ramdane-Cherif.
\newblock {Competitive memetic algorithms for arc routing problems}.
\newblock {\em Annals of Operations Research}, 131(1):159--185, 2004.

\bibitem{longo2006solving}
H.~Longo, M.P. de~Arag{\~a}o, and E.~Uchoa.
\newblock {Solving capacitated arc routing problems using a transformation to
  the CVRP}.
\newblock {\em Computers and Operations Research}, 33(6):1823--1837, 2006.

\bibitem{martinelli2011improved}
R.~Martinelli, M.~Poggi, and A.~Subramanian.
\newblock {Improved bounds for large scale capacitated arc routing problem}.
\newblock {\em Computers $\&$ Operations Research}, 40(8):2145--2160, 2013.

\bibitem{mei2013decomposing}
Y.~Mei, X.~Li, and X.~Yao.
\newblock {Decomposing large-scale capacitated arc routing problems using a
  random route grouping method}.
\newblock In {\em Proceedings of the 2013 IEEE Congress on Evolutionary
  Computation}, pages 1013--1020, 2013.

\bibitem{mei2014cooperative}
Y.~Mei, X.~Li, and X.~Yao.
\newblock {Cooperative coevolution with route distance grouping for large-scale
  capacitated arc routing problems}.
\newblock {\em IEEE Transactions on Evolutionary Computation}, 18(3):435--449,
  2014.

\bibitem{mei2009global}
Y.~Mei, K.~Tang, and X.~Yao.
\newblock {A global repair operator for capacitated arc routing problem}.
\newblock {\em IEEE Transactions on Systems, Man, and Cybernetics, Part B:
  Cybernetics}, 39(3):723--734, 2009.

\bibitem{mei2011memetic}
Y.~Mei, K.~Tang, and X.~Yao.
\newblock {A memetic algorithm for periodic capacitated arc routing problem}.
\newblock {\em IEEE Transactions on Systems, Man, and Cybernetics, Part B:
  Cybernetics}, 41(6):1654--1667, 2011.

\bibitem{mei2011decomposition}
Y.~Mei, K.~Tang, and X.~Yao.
\newblock {Decomposition-based memetic algorithm for multiobjective capacitated
  arc routing problem}.
\newblock {\em IEEE Transactions on Evolutionary Computation}, 15(2):151--165,
  2011.

\bibitem{mourao2005heuristic}
M.C. Mour{\~a}o and L.~Amado.
\newblock {Heuristic method for a mixed capacitated arc routing problem: A
  refuse collection application}.
\newblock {\em European Journal of Operational Research}, 160(1):139--153,
  2005.

\bibitem{su2014automatic}
S.~Nguyen, M.J. Zhang, M.~Johnston, and K.C. Tan.
\newblock {Automatic design of scheduling policies for dynamic multi-objective
  job shop scheduling via cooperative coevolution genetic programming}.
\newblock {\em IEEE Transactions on Evolutionary Computation}, 18(2):193--208,
  2014.

\bibitem{oliveira2016cooperative}
F.B.D. Oliveira, R.~Enayatifar, H.J. Sadaei, and J.Y. Potvin.
\newblock {A cooperative coevolutionary algorithm for the multi-depot vehicle
  routing problem}.
\newblock {\em Expert Systems with Applications}, 43(C):117--130, 2016.

\bibitem{ostertag2009POPMUSIC}
A.~Ostertag, K.F. Doerner, R.F. Hartl, E.D. Taillard, and P.~Waelti.
\newblock Popmusic for a real-world large-scale vehicle routing problem with
  time windows.
\newblock {\em Journal of the Operational Research Society}, 60(7):934--943,
  2009.

\bibitem{polacek2008variable}
M.~Polacek, K.F. Doerner, R.F. Hartl, and V.~Maniezzo.
\newblock {A variable neighborhood search for the capacitated arc routing
  problem with intermediate facilities}.
\newblock {\em Journal of Heuristics}, 14(5):405--423, 2008.

\bibitem{shang2016improved}
R.~Shang, K.~Dai, L.~Jiao, and et~al.
\newblock {Improved memetic algorithm based on route distance grouping for
  multiobjective large scale capacitated arc routing problems}.
\newblock {\em IEEE Transactions on Cybernetics}, 46(4):1000--1013, 2016.

\bibitem{shang2017memetic}
R.~Shang, B.~Du, K.~Dai, and et~al.
\newblock {Memetic algorithm based on extension step and statistical filtering
  for large--scale capacitated arc routing problems}.
\newblock {\em Natural Computing}, pages 1--17, 2017.

\bibitem{shang2017quantum-Inspired}
R.~Shang, B.~Du, K.~Dai, L.~Jiao, and et~al.
\newblock {Quantum--inspired immune clonal algorithm for solving large--scale
  capacitated arc routing problems}.
\newblock {\em Memetic Computing}, pages 1--22, 2017.

\bibitem{shang2016immune}
R.H. Shang, H.N. Ma, J.~Wang, L.C. Jiao, and R.~Stolkin.
\newblock {Immune clonal selection algorithm for capacitated arc routing
  problem}.
\newblock {\em Soft Computing}, 20(6):2177--2204, 2016.

\bibitem{tan2006distributed}
K.C. Tan, Y.J. Yang, and C.K. Goh.
\newblock {A distributed cooperative coevolutionary algorithm for
  multiobjective optimization}.
\newblock {\em IEEE Transactions on Evolutionary Computation}, 10(5):527--549,
  2006.

\bibitem{tang2009memetic}
K.~Tang, Y.~Mei, and X.~Yao.
\newblock {Memetic algorithm with extended neighborhood search for capacitated
  arc routing problems}.
\newblock {\em IEEE Transactions on Evolutionary Computation},
  13(5):1151--1166, 2009.

\bibitem{tang2017scalable}
K.~Tang, J.~Wang, X.~Li, and X.~Yao.
\newblock {A scalable approach to capacitated arc routing problems based on
  hierarchical decomposition}.
\newblock {\em IEEE Transactions on Cybernetics}, 47(11):3928--3940, 2017.

\bibitem{ulusoy1985fleet}
G.~Ulusoy.
\newblock {The fleet size and mix problem for capacitated arc routing}.
\newblock {\em European Journal of Operational Research}, 22(3):329--337, 1985.

\bibitem{vidal2017node}
T.~Vidal.
\newblock {Node, edge, arc routing and turn penalties: Multiple problems-one
  neighborhood extension}.
\newblock {\em Operations Research}, 65(4):992--1010, 2017.

\bibitem{wilcoxon1945individual}
F.~Wilcoxon.
\newblock {Individual comparisons by ranking methods}.
\newblock {\em Biometrics Bulletin}, 1(6):80--83, 1945.

\bibitem{wohlk2018fast}
S.~W{\o}hlk and G.~Laporte.
\newblock {A fast heuristic for large-scale capacitated arc routing problem}.
\newblock {\em Journal of the Operational Research Society}, 69(12):1877--1887,
  2018.

\bibitem{xing2010evolutionary}
L.N. Xing, P.~Rohlfshagen, Y.W. Chen, and X.~Yao.
\newblock {An evolutionary approach to the multidepot capacitated arc routing
  problem}.
\newblock {\em IEEE Transactions on Evolutionary Computation}, 14(3):356--374,
  2010.

\bibitem{yang2008large}
Z.Y. Yang, K.~Tang, and X.~Yao.
\newblock {Large scale evolutionary optimization using cooperative
  coevolution}.
\newblock {\em Information Sciences}, 178(15):2985--2999, 2008.

\bibitem{yao2017memetic}
T.T. Yao, X.~Yao, S.S. Han, Y.C. Wang, D.P. Cao, and F.Y. Wang.
\newblock {Memetic algorithm with adaptive local search for capacitated arc
  routing problem}.
\newblock In {\em IEEE 20th International Conference on Intelligent
  Transportation Systems}, pages 1--6, 2017.

\bibitem{zbib2017vriants}
H.~Zbib.
\newblock {Variants of the path scanning construction heuristic for the
  no-split multi-compartment capacitated arc routing problem(Working Paper)}.
\newblock {\em Aarhus University}, 2017.

\bibitem{zhang2010divide-and-conquer}
R.~Zhang and C.~Wu.
\newblock {A divide-and-conquer strategy with particle swarm optimization for
  the job shop scheduling problem}.
\newblock {\em Engineering Optimization}, 42(7):641--670, 2010.

\bibitem{zhang2017memetic}
Y.Z. Zhang, Y.~Mei, K.~Tang, and K.Q. Jiang.
\newblock {Memetic algorithm with route decomposing for periodic capacitated
  arc routing problem}.
\newblock {\em Applied Soft Computing}, 52(3):1130--1142, 2017.

\end{thebibliography}

\end{document}